\documentclass[number,preprint,review,12pt]{elsarticle}

\usepackage{amssymb}
\usepackage{amsmath}
\usepackage{xcolor}

\usepackage{tabularx} 
\usepackage{booktabs} 
\usepackage{multirow} 
\usepackage{graphicx}
\usepackage{subcaption}
\usepackage{hyperref}
\usepackage{bibunits}
\defaultbibliographystyle{elsarticle-num}
\defaultbibliography{refs}      

\usepackage[a4paper,margin=3cm]{geometry}

\journal{Electric Power Systems Research}

\begin{document}

\begin{bibunit}

\begin{frontmatter}



\title{LOAD FORECASTING ON A HIGHLY SPARSE ELECTRICAL LOAD DATASET USING GAUSSIAN INTERPOLATION}


\author[add1]{Chinmoy Biswas\fnref{fn1}} 
\address[add1]{Department of Electrical and Electronic Engineering, Bangladesh University of Engineering and Technology (BUET), Dhaka-1205, Bangladesh}
\author[add1]{Nafis Faisal\fnref{fn1}}
\author[add1]{Vivek Chowdhury}
\author[add1]{Abrar Al-Shadid Abir}
\author[add1]{Sabir Mahmud}
\author[add1]{Mithon Rahman}
\author[add1]{Shaikh Anowarul Fattah}
\author[add1]{Hafiz Imtiaz\corref{cor1}}
\ead{hafizimtiaz@eee.buet.ac.bd}

\cortext[cor1]{Corresponding author}
\fntext[fn1]{These authors contributed equally}

\begin{abstract}
Sparsity, defined as the presence of missing or zero values in a dataset, often poses a major challenge while operating on real-life datasets. Sparsity in features or target data of the training dataset can be handled using various interpolation methods, such as linear or polynomial interpolation, spline, moving average, or can be simply imputed. Interpolation methods usually perform well with Strict Sense Stationary (SSS) data. In this study, we show that an approximately 62\% sparse dataset with hourly load data of a power plant can be utilized for load forecasting assuming the data is Wide Sense Stationary (WSS), if augmented with Gaussian interpolation. More specifically, we perform statistical analysis on the data, and train multiple machine learning and deep learning models on the dataset. By comparing the performance of these models, we empirically demonstrate that Gaussian interpolation is a suitable option for dealing with load forecasting problems. Additionally, we demonstrate that Long Short-term Memory (LSTM)-based neural network model offers the best performance among a diverse set of classical and neural network-based models.
\end{abstract}




\end{frontmatter}




\section{Introduction}
\label{sec1}

Missing data in either feature or target values of a dataset, also known as sparsity, is a major problem in practical machine learning (ML) model training~\cite{palanivinayagam2023missing}. According to the manifold hypothesis~\cite{gorban2018blessing, fefferman2016testing} by considering data points as vectors in a high dimensional space, training a model is essentially fitting to the geometry of the lower dimensional manifold inside the data space that contains all the data. Having missing data means incomplete information about the data manifold. This causes the model to fit improperly to the geometry of the manifold. Due to sparsity being a significant challenge in real-life datasets, the study of the reconstruction of a signal from a small number of samples in a domain in which the signal is sparse, or \emph{compressed sensing}, is an active field of research~\cite{donoho2006compressed, candes2006robust}. 

Precise forecasting of the electric load demand is important for the economic operation of power generation systems \cite{bunn1982short, xu2025deep}. Inaccurate forecasts may lead to an increase in the operating cost of power plants, and complete blackout~\cite{haida2002regression, dehbozorgi2025deep}. Several factors, such as time, weather, resident’s behavior, may affect load demand at a particular area. Based on time range, load forecasting can be broadly divided into three categories: short-term forecasts (one hour to one-week) \cite{he2025short}, medium forecasts (one week to one year) and long-term forecasts (more than a year) \cite{Singh2012LoadFT}. \\

\noindent\textbf{Related Works. }In general, load forecasting methods can be divided into two types: classical methods and deep learning-based techniques. Classical methods use statistical approaches to forecast the demand based on the previous records of numerical data. Many classical methods of load forecasting have been introduced in the past few decades, such as regression \cite{Papalexopoulos1989regression}, exponential smoothing \cite{Christiaanse1971Short}, Autoregressive Moving Average (ARMA) \cite{Timothy1995neural} and Box-Jenkins models \cite{sadaei2019short}. However, these methods are shown to be inefficient in establishing a precise or economic forecasting model because of the complicated and non-linear nature of electrical load time series. As such, deep learning-based techniques are introduced to overcome the limitations of the traditional approaches \cite{priyadarsini2025cnn}. To this end, Recurrent Neural Network (RNN) \cite{Aseeri2023effective}, Gated Recurrent Unit (GRU) \cite{Li2022research}, Long Short-Term Memory (LSTM) \cite{Pengdan2024novel, tulensalo2020lstm} are some widely used deep learning models in the field of load forecasting \cite{rosseel2025physics}.

Among existing methods of interpolation for missing data-point imputation linear \cite{huang2021missing}, polynomial \cite{noor2014filling}, moving average \cite{thompson1952construction}, Kalman filtering \cite{gomez1994estimation}, exponential smoothing \cite{kai2008nurbs}, splines \cite{wahba1981spline}, piecewise cubic Hermite \cite{rabbath2019comparison} are some commonly used examples. However, these methods usually work well with datasets having smaller proportion of missing data. Additionally, these approaches do not perform well on datasets that can be modeled as Wide Sense Stationary (WSS)~\cite{tjostheim1975some} and non-stationary random processes \cite{priestley1967power}, since the statistical features are not consistent with time \cite{edition2002probability}. Sparsity in such datasets is a major challenge, and having over 50\% missing data complicates training an ML model on such dataset further. Although some techniques, such as differencing \cite{mills2011dealing} and distending \cite{dombi2020new}, can be used to overcome the non-stationarity to some extent, sparsity and non-stationarity pose significant hurdles for training ML models in general~\cite{poulinakis2023machine}.\\

\noindent\textbf{Our Contribution. }In this work, we intend to address this research gap. More specifically, we consider the problem of load forecasting from a very sparse dataset by interpolating the missing data in the time domain, assuming the data generation is a WSS process. We use the interpolated dataset to train several off-the-shelf machine learning models for load forecasting: LSTM~\cite{Singh2012LoadFT}, BiLSTM (Bidirectional LSTM)~\cite{huang2015bidirectional}, CNN-LSTM~\cite{lu2020cnn}, CNN-BiLSTM~\cite{lu2021cnn}, Transformer\cite{vaswani2017attention}, Prophet\cite{jha2021time}, DLinear (Dynamic Linear)~\cite{zeng2023transformers}, SARIMA (Seasonal AutoRegressive Integrated Moving Average)~\cite{Chen2018time}, XGBoost~\cite{chen2016xgboost}, Random Forest~\cite{rigatti2017random}. We empirically demonstrate that among these models, the LSTM-based model offers the best performance.

\section{Proposed Methodology}
\label{sec2}
We reviewed the neural network models that are necessary for our work in the supplementary section~\ref{app:background}. In the following, we describe our proposed methodology in detail.

\subsection{Data Collection}
\label{subsec1}

The dataset that we primarily utilize in this work is the daily load demand of Bangladesh University of Engineering and Technology (BUET) campus, including its residential dormitory, from 2015, 2016, 2021 and 2022. In most cases, the load demand of BUET is primarily fulfilled from its own power plant from 8 AM to 11 PM (16 Hours). Due to the COVID-19 pandemic and poor condition of the manually collected data, the data of 2017--2020 could not be utilized. In addition to the load demand data, we included temperature, wind speed, relative humidity, and precipitation information to our dataset for capturing the seasonal and atmospheric effects. These additional information were collected from the NASA Prediction Of Worldwide Energy Resources (POWER) website \cite{nasapower}. Bangladesh holiday information is also added. Considering 16 hours per day, the sparsity of this dataset is approximately 62.45\%. For benchmarking the results, the Dayton Electrical Load Demand from 2004 to 2018 dataset from the PJM Hourly Energy Consumption data bank \cite{pjmdataset} is used.

\begin{figure}[t]
\centering
\includegraphics[width=0.8\textwidth]{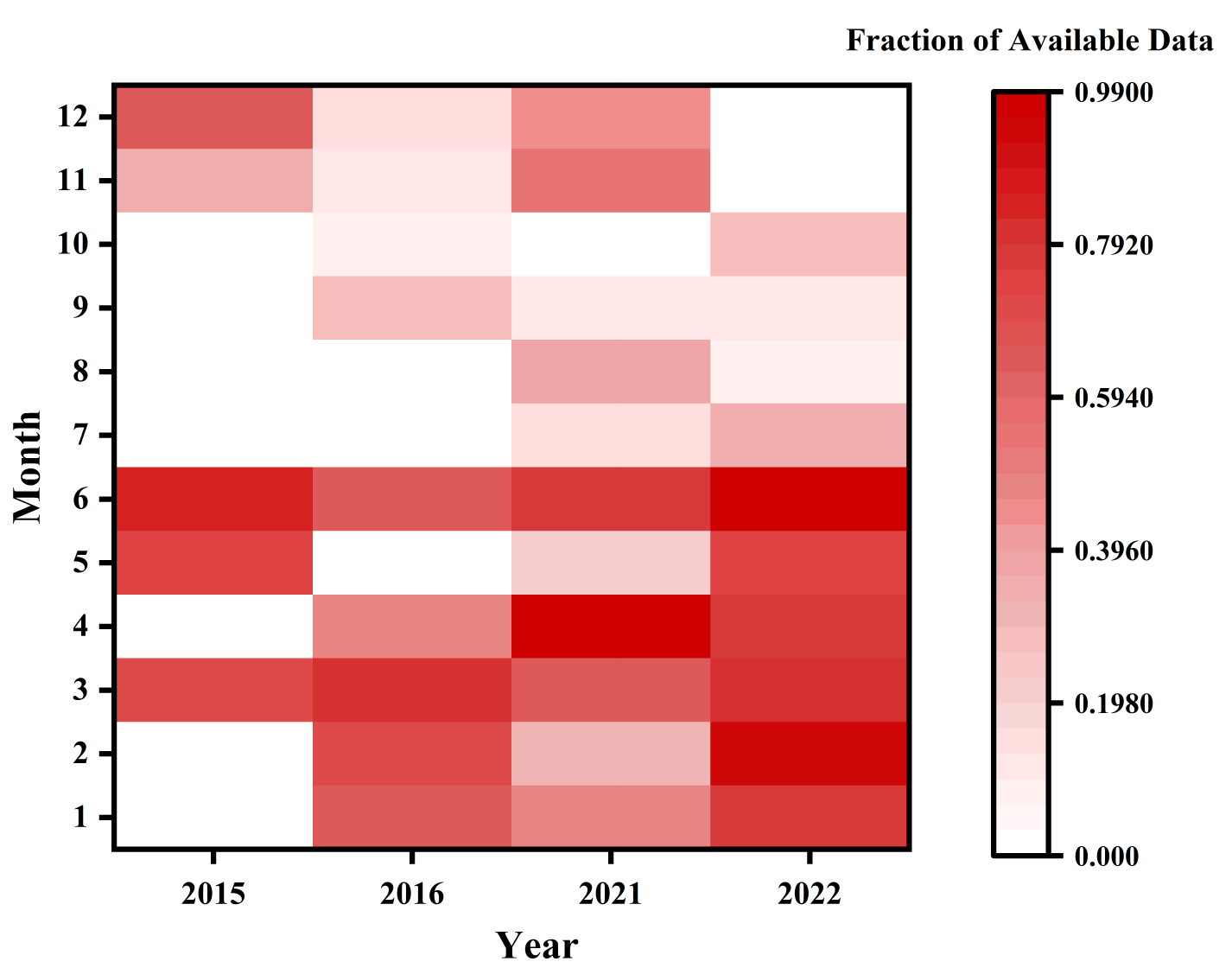}
\caption{Heat map of missing data by month and year}
\label{heatmap}
\end{figure}

\subsection{Interpolation of Missing Data}
\label{subsec2}
As mentioned before, more than 60\% data is missing in the BUET power plant load demand dataset. In Fig.~\ref{heatmap}, we show a heat map of missing data by year and month. Due to the extent of missing data, commonly used interpolation methods, such as linear, polynomial or spline interpolation, could not be used for a good load forecasting performance. To address this, we propose to impute the missing data-points by estimating a Gaussian probability density function (PDF) from the existing data-points, and then sampling from the PDF. A Gaussian PDF with mean $\mu$ and variance $\sigma^2$ can be defined as \cite{edition2002probability},
\begin{equation}
    f_X(x) = \frac{1}{\sqrt{2\pi\sigma^2}} \exp\left(-\frac{(x - \mu)^2}{2\sigma^2}\right).
\end{equation}
Note that, from a standard Normal random variable $Z \sim \mathcal{N}(0, 1)$, $X$ can be defined as $X = \mu + \sigma Z$. The latter is the approach used in this paper. We note that the BUET power plant's maximum generating capacity was 2000~kW during the time we are considering. Therefore, any random sample with value above 2000~kW were clipped. Similarly, negative samples were clipped to zero. We emphasize that the Gaussian PDF for each hour is different with different $(\mu,\sigma)$, since we assume that the load data is WSS.

\begin{figure}[t]
\includegraphics[width=1\textwidth]{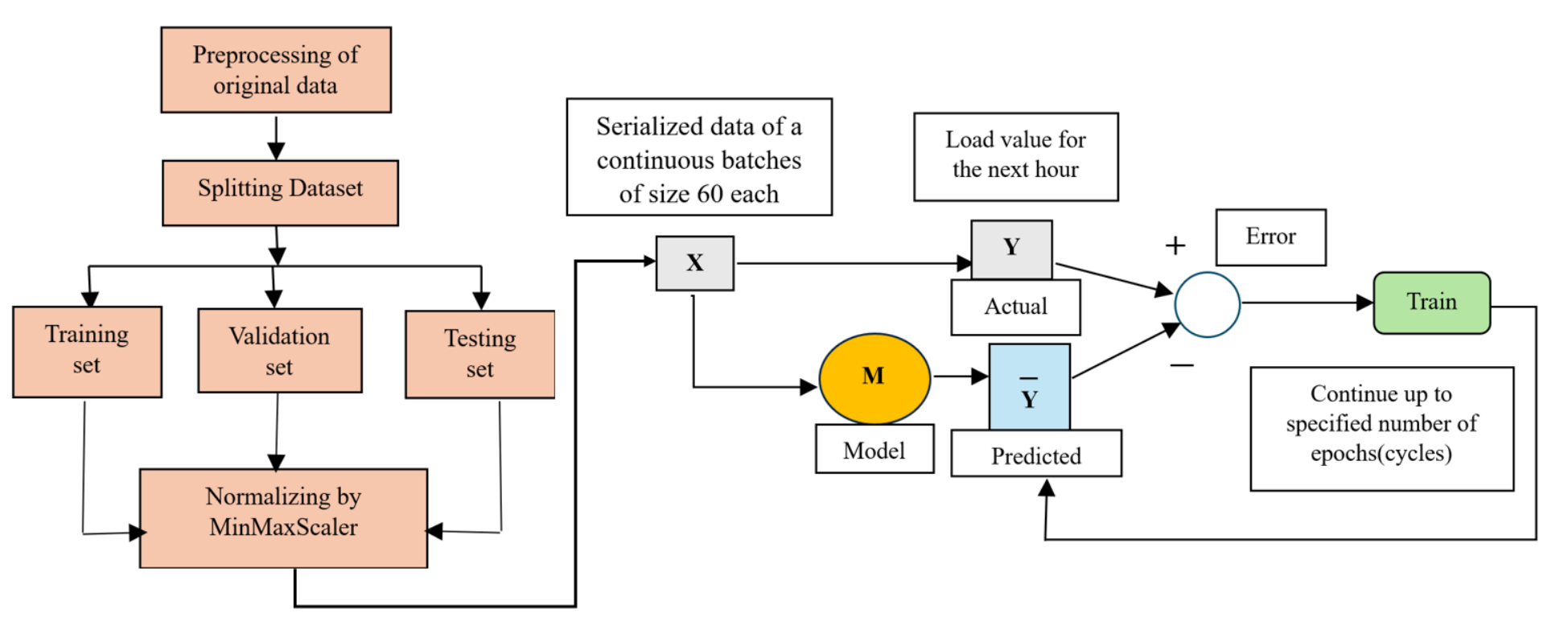}
\caption{Schematic of model training for RNN-based models}
\label{workflow}
\end{figure}

\subsection{Model Training}
\label{subsec3}
To obtain load forecasts from the interpolated dataset, we trained multiple predictive models. The selected features included Year, Month, Day, Weekday, Hour, average temperature, wind speed, relative humidity, and the target variable, Total Load. Prior to training, all dataset values were normalized using \texttt{MinMaxScaler}~\cite{priyambudi2024algorithm}. The dataset was partitioned such that 66\% of the samples were used for training, while the remaining portion were reserved for validation and testing purposes. The models employed encompassed a range of methods from classical statistical models like SARIMA, advanced machine learning algorithms such as XGBoost, Random Forest, DLinear, and deep-learning-based models including LSTM, BiLSTM, CNN-LSTM, CNN-BiLSTM, Transformer, and Prophet.

For the RNN-based models, the data was structured into sequential windows comprising features and target values of a specific length (denoted as sequence length), which served as inputs to predict the target values of the subsequent window of specified length (denoted as prediction length). Given that the input for time series forecasting models consists of historical data from previous time frames, the dataset was serialized so that hourly load, weather, and time features within a defined window were used to predict the load demand for the following hour. The serializing window was shifted by one hour iteratively until the entire dataset was exhausted. Fig.~\ref{workflow} illustrates the training workflow for the RNN-based models. Additionally, the Transformer model required positional embeddings, which were provided using sine and cosine positional encoders.

In contrast, Prophet, SARIMA, Random Forest, and XGBoost models do not require data serialization. More specifically, Prophet and SARIMA require only the Total Load target variable without additional features. The Prophet model accommodated holidays as supplementary input data, whereas SARIMA did not require any additional parameters. Both Random Forest and XGBoost required feature extraction from the target column, with extracted features serving as input and load values as output.

Finally, the model demonstrating superior performance was further trained and validated on the Dayton dataset. This facilitated a comparative analysis of prediction accuracy between the augmented dataset and a real-world dataset.

\subsection{Global Hyper-parameters and Configurations}
\label{subsec4}

The models were trained using sequences of length 64 to predict one time step ahead into the future. The training batch size was 32 for 40 epochs using the Adam optimizer with an initial learning rate of 0.0001. A learning rate scheduler was used to reduce the learning rate by a factor of 0.5 on plateau of the validation loss, with patience of 3 epochs. Early stopping was also used, monitoring validation loss with patience of 10 epochs, to prevent overfitting. Dropout regularization was 0.2 wherever applicable in the network architecture. The Mean Squared Error (MSE) was used as the loss function and evaluation metric. The model that performed best was saved as the version that had the lowest validation loss. A list of hyperparameters and model settings is given in Table~\ref{table1}.

\begin{table*}[t]
\centering
\caption{Configurations and hyper-parameters}
\begin{tabularx}{\textwidth}{@{} l X @{}}
\toprule
\textbf{Model Name} & \textbf{Model Specific hyper-parameters} \\ \midrule
LSTM               & Hidden layers = 128 \\ \hline
BiLSTM             & Hidden layers = 128 \\ \hline
CNN-LSTM           & Conv layers = 2, Kernel sizes = 64, 128; Hidden layers = 128 \\ \hline
CNN-BiLSTM         & Conv layers = 2, Kernel sizes = 64, 128; Hidden layers = 128 \\ \hline
Transformer        & Embedding space dimension = 64; Attention heads = 4; Number of layers = 12; Sine and Cosine for positional encoding \\ \hline
Prophet            & Holidays dataframe, other parameters default \\ \hline
DLinear            & Kernel size = 25; Individual models for each feature set to False \\ \hline
SARIMA             & $(p, d, q) = (5, 1, 1)$ \\ \hline
Random Forest      & Number of trees in parallel or estimators = 2000; Number of cores used = 4; Maximum depth of each tree = 200 \\ \hline
XGBoost            & Number of trees in ensemble or estimators = 2000; Number of cores used = 6; Maximum depth of each tree = 200 \\ \bottomrule
\end{tabularx}
\label{table1}
\end{table*}

\section{Results}
\label{sec3}

\subsection{Statistical Analyses on the Interpolated Dataset}
\label{subsec5}

\begin{figure}[t]
    \centering
    \begin{minipage}{0.49\textwidth}
    \centering
    \includegraphics[width=1\textwidth]{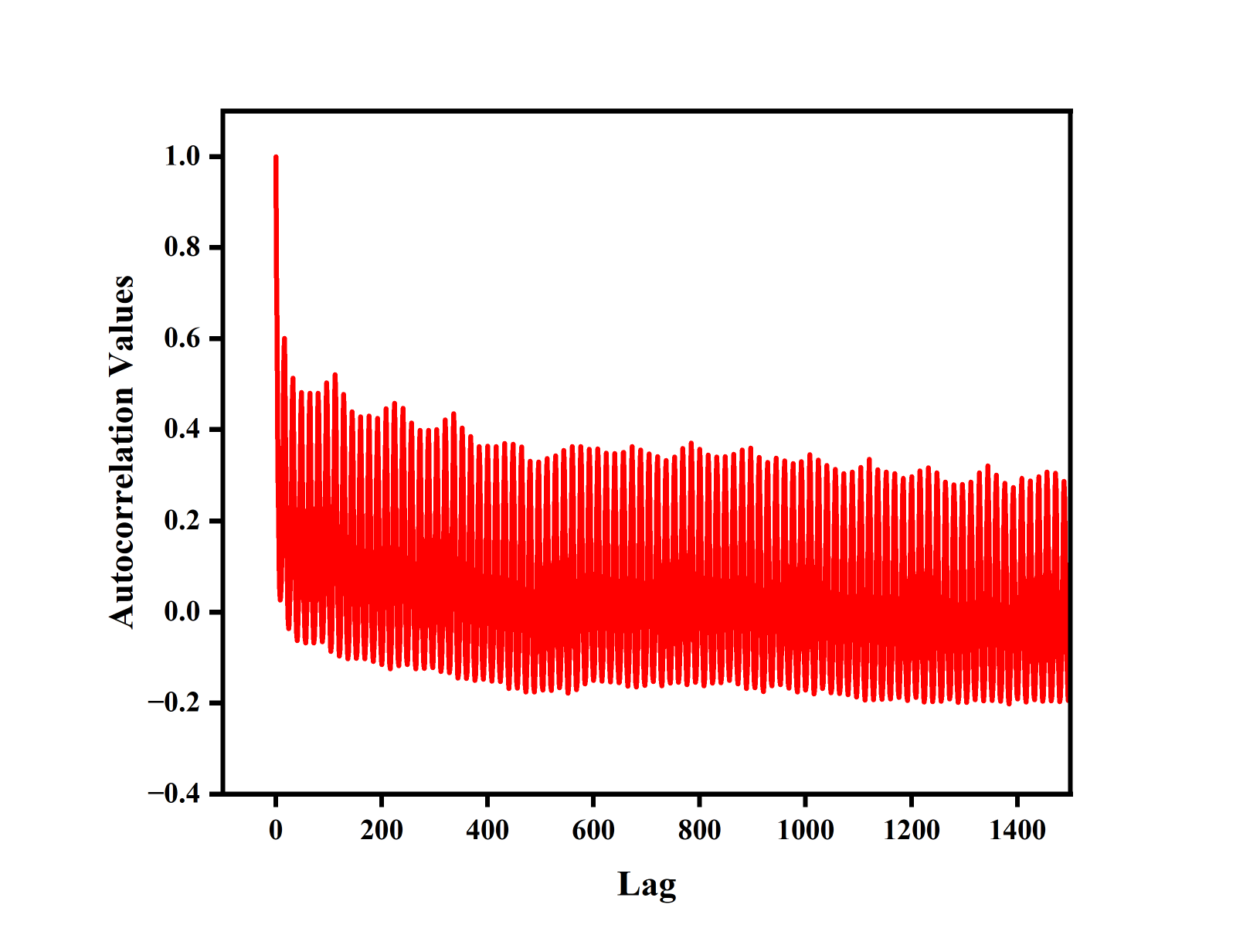}
    \caption{Autocorrelation of the Interpolated Data}
    \label{ACC}
    \end{minipage}
    \hfill
    \begin{minipage}{0.49\textwidth}
    \centering
    \includegraphics[width=1\textwidth]{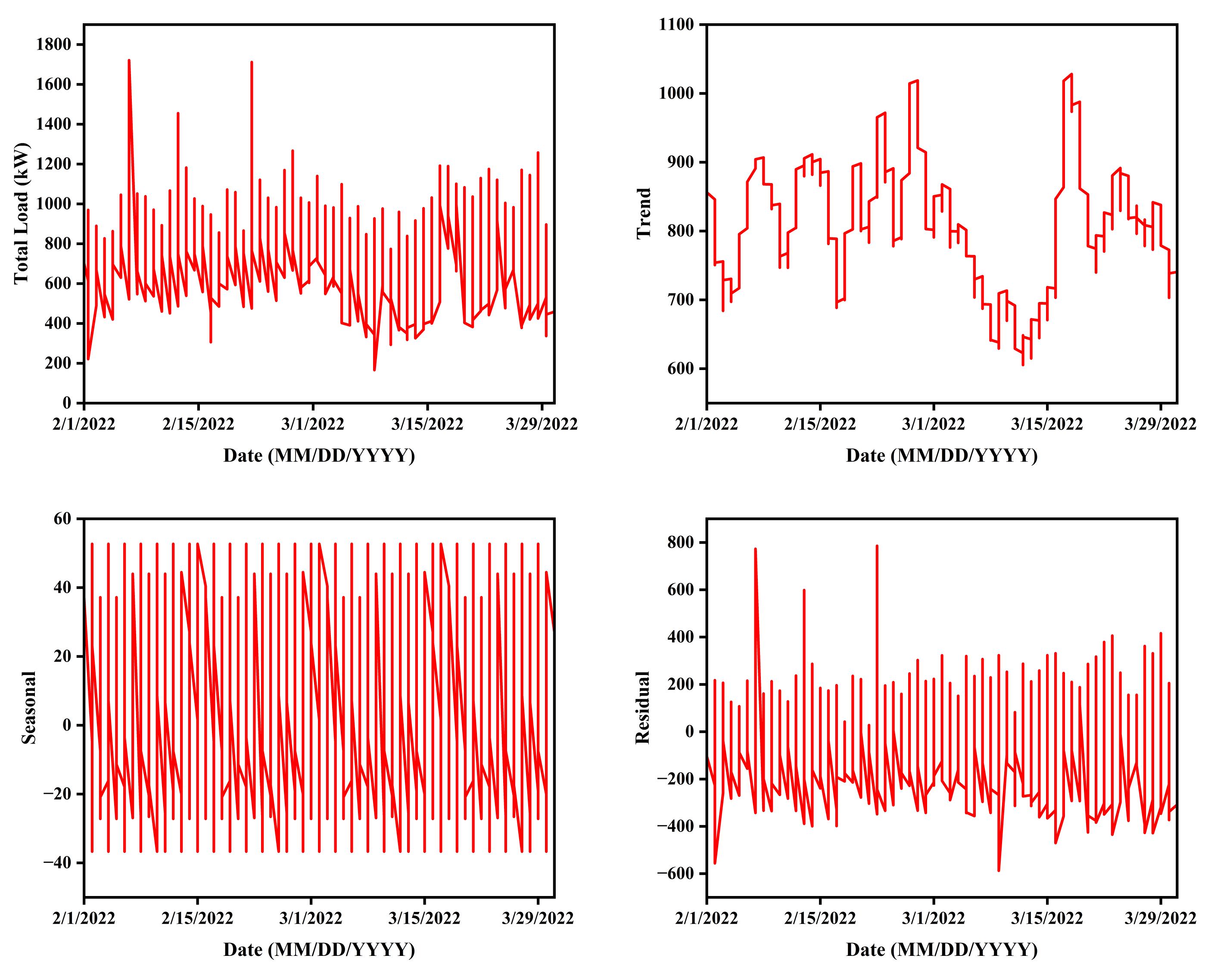}
    \caption{Additive time series decomposition of the interpolated data}
    \label{ATSD}
    \end{minipage}   
\end{figure}

The initial statistical assessment for any time-series dataset involves verifying whether the dataset exhibits characteristics of white noise. This was evaluated by computing the auto-correlation~\cite{bence1995analysis} of the time-series data, as illustrated in Fig.~\ref{ACC}. The results indicate that the dataset maintains high correlation values across different lag parameters, in addition to the zero-lag case. This suggests that the dataset contains meaningful temporal dependencies, making it suitable for time-series prediction. Additionally, the auto-correlation remains notably high for extended lags, which can be attributed to the Gaussian interpolation applied to address missing data.

To analyze the seasonality and trend characteristics of the dataset, an additive time series decomposition \cite{xu2019temporal} was performed, as illustrated in Fig.~\ref{ATSD}. The decomposition revealed a highly non-stationary trend component, which is expected given the dataset's sparse nature, interpolation process, and that it is resulting from a WSS process.

To further assess the stationarity of the dataset, the Augmented Dickey-Fuller (ADF) test \cite{dickey1979distribution} and the Kwiatkowski–Phillips–Schmidt–Shin (KPSS) test \cite{kwiatkowski1992testing} were conducted. Both tests returned $p$-values smaller than 0.05. For the ADF test, this result indicates that the null hypothesis, which assumes the presence of a unit root and thus non-stationarity, can be rejected, confirming stationarity. On the other hand, the KPSS test suggests that the null hypothesis of stationarity can be rejected, indicating that the dataset is non-stationary \cite{box2015time}. These results align with expectations for a dataset exhibiting WSS properties. Although various techniques, such as data lagging \cite{Ng01031995}, are commonly employed to address seasonality, we observed that such approaches introduce significant forecasting errors when applied to a WSS-interpolated dataset. As a result, these methods were excluded from further analysis.

\begin{figure}[t]
    \centering
    \begin{subfigure}[t]{0.32\textwidth}
        \centering
        \includegraphics[width=1\textwidth]{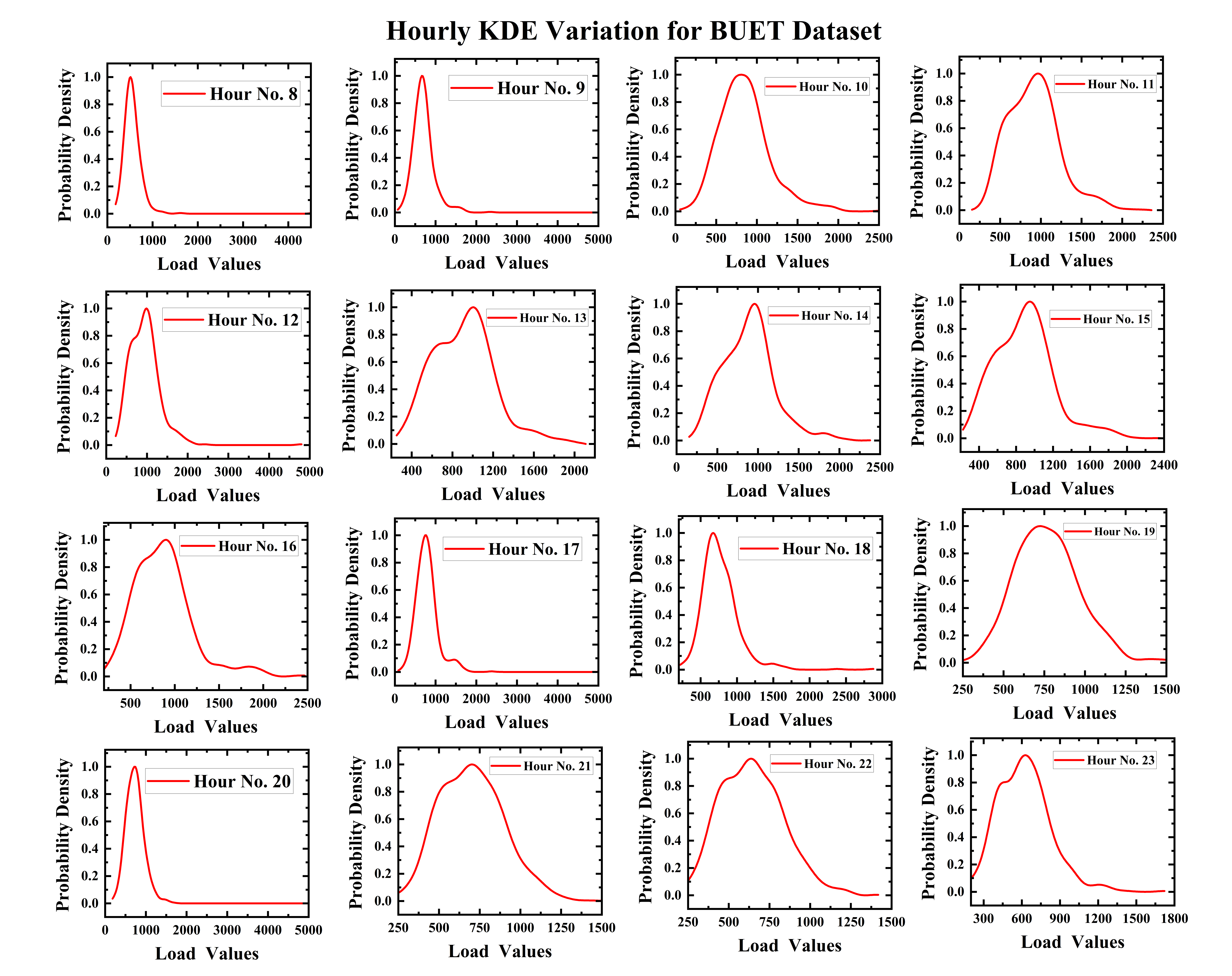}
        \caption{Hourly}
        \label{BUET_hourly_KDE}
    \end{subfigure}
    \hfill
    \begin{subfigure}[t]{0.32\textwidth}
        \centering
        \includegraphics[width=1\textwidth]{images/BUET_daily_KDE.pdf}
        \caption{Daily}
        \label{BUET_daily_KDE}
    \end{subfigure}
    \hfill
    \begin{subfigure}[t]{0.32\textwidth}
        \centering
        \includegraphics[width=1\textwidth]{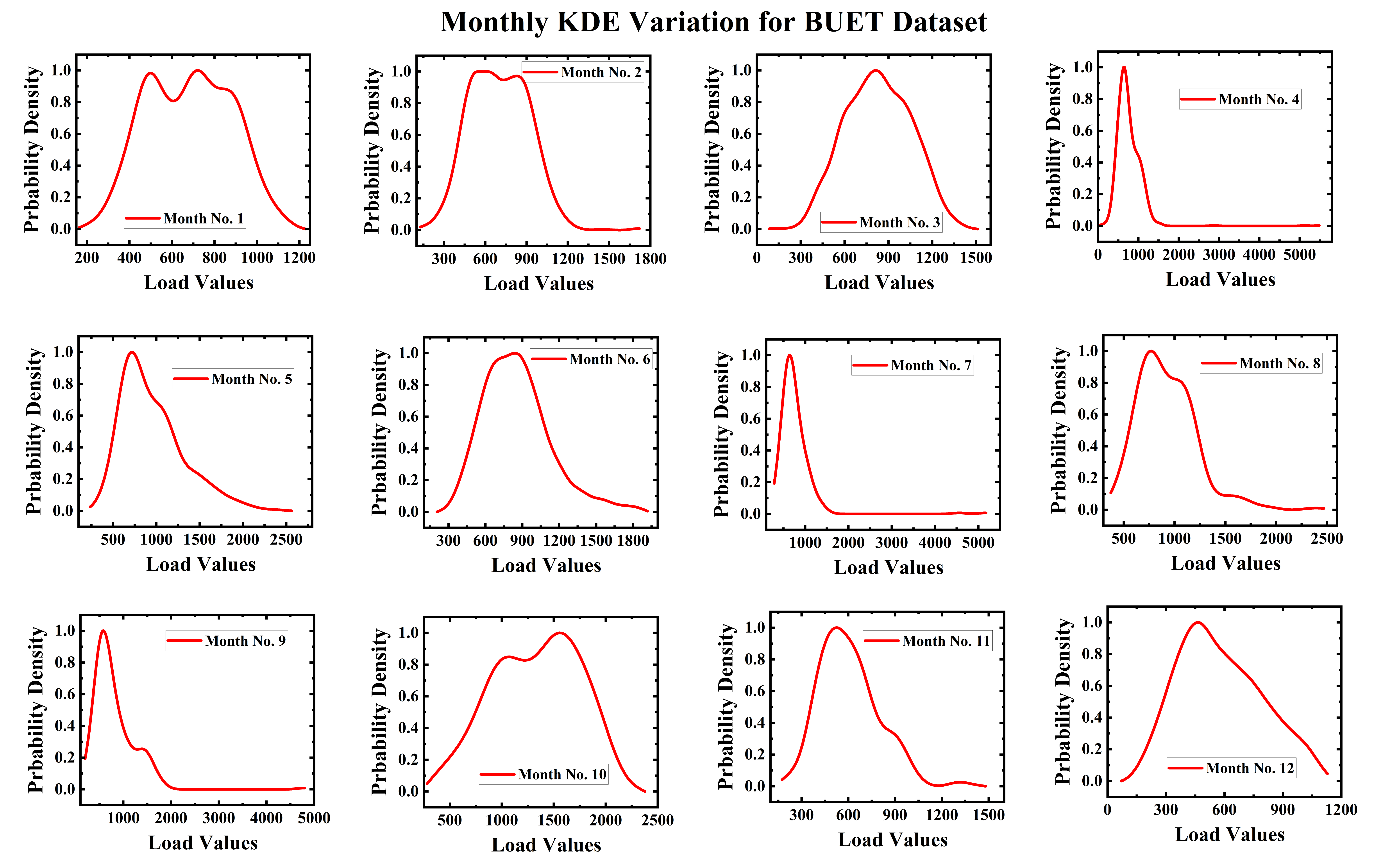}
        \caption{Monthly}
        \label{BUET_monthly_KDE}
    \end{subfigure}
    \caption{Load probability densities}
    \label{Load_Probability_Densities}
\end{figure}

The hourly, daily, and monthly Kernel Density Estimation (KDE) \cite{dubnicka2009kernel} of the dataset (including missing values) were analyzed for a non-parametric estimate of the probability density function (PDF) of the load random variable. Fig.~\ref{BUET_hourly_KDE}, Fig.~\ref{BUET_daily_KDE}, and Fig.~\ref{BUET_monthly_KDE} illustrate the hourly, daily, and monthly KDEs of the dataset, respectively. The horizontal axis represents the load bin values, while the vertical axis denotes the probability density. To enhance visualization, the probability densities are scaled by their respective maximum values. The resulting PDFs exhibit a bell-shaped curve but do not follow a consistent pattern across different time frames. This observation motivated the exploration of multiple distribution functions (DF) for interpolation.

\begin{figure}[t]
    \centering
    \begin{subfigure}[t]{0.32\textwidth}
        \centering
        \includegraphics[width=1\textwidth]{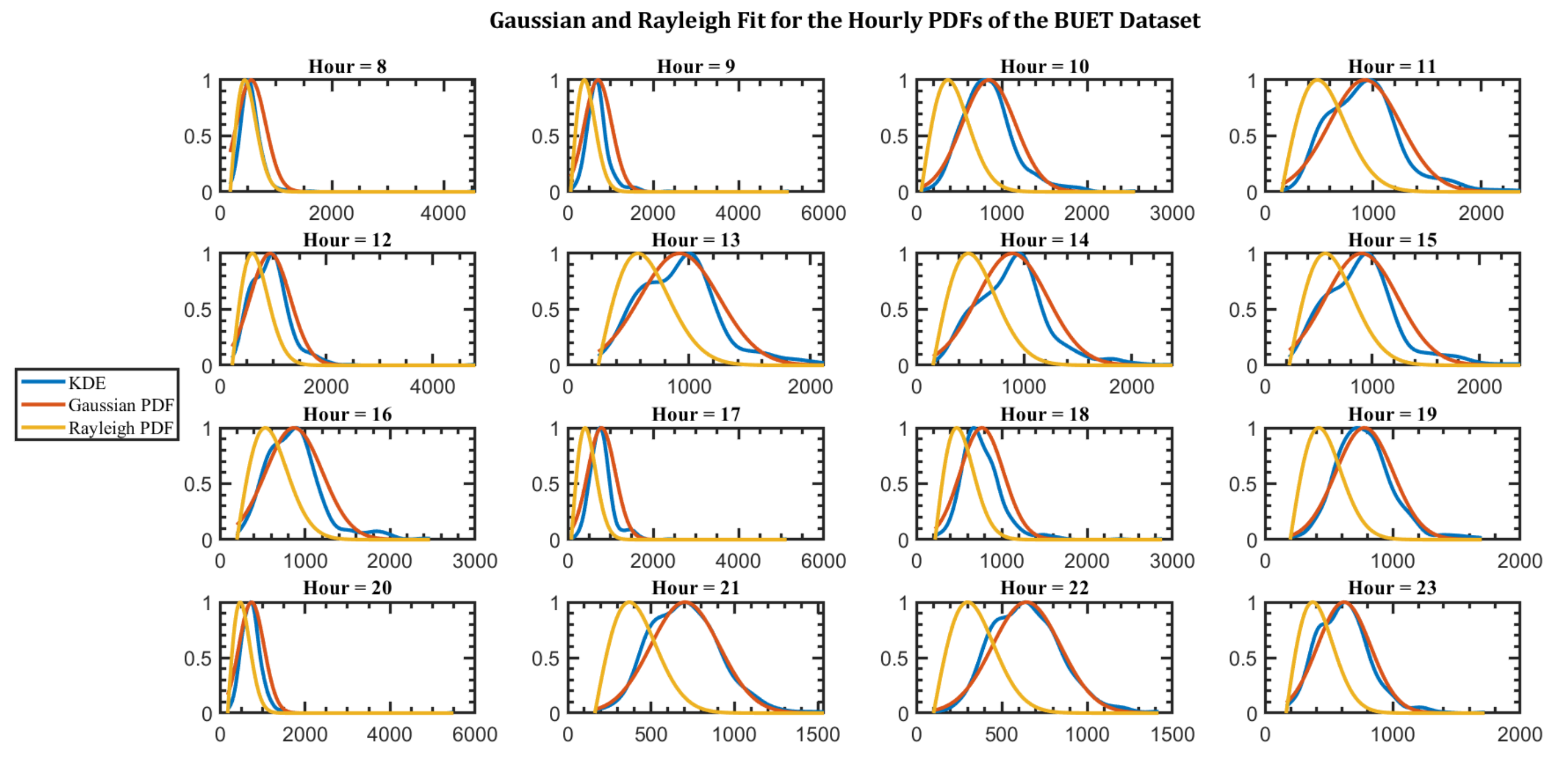}
        \caption{Hourly}
        \label{BUET_hourly_dist}
    \end{subfigure}
    \hfill
    \begin{subfigure}[t]{0.32\textwidth}
        \centering
        \includegraphics[width=1\textwidth]{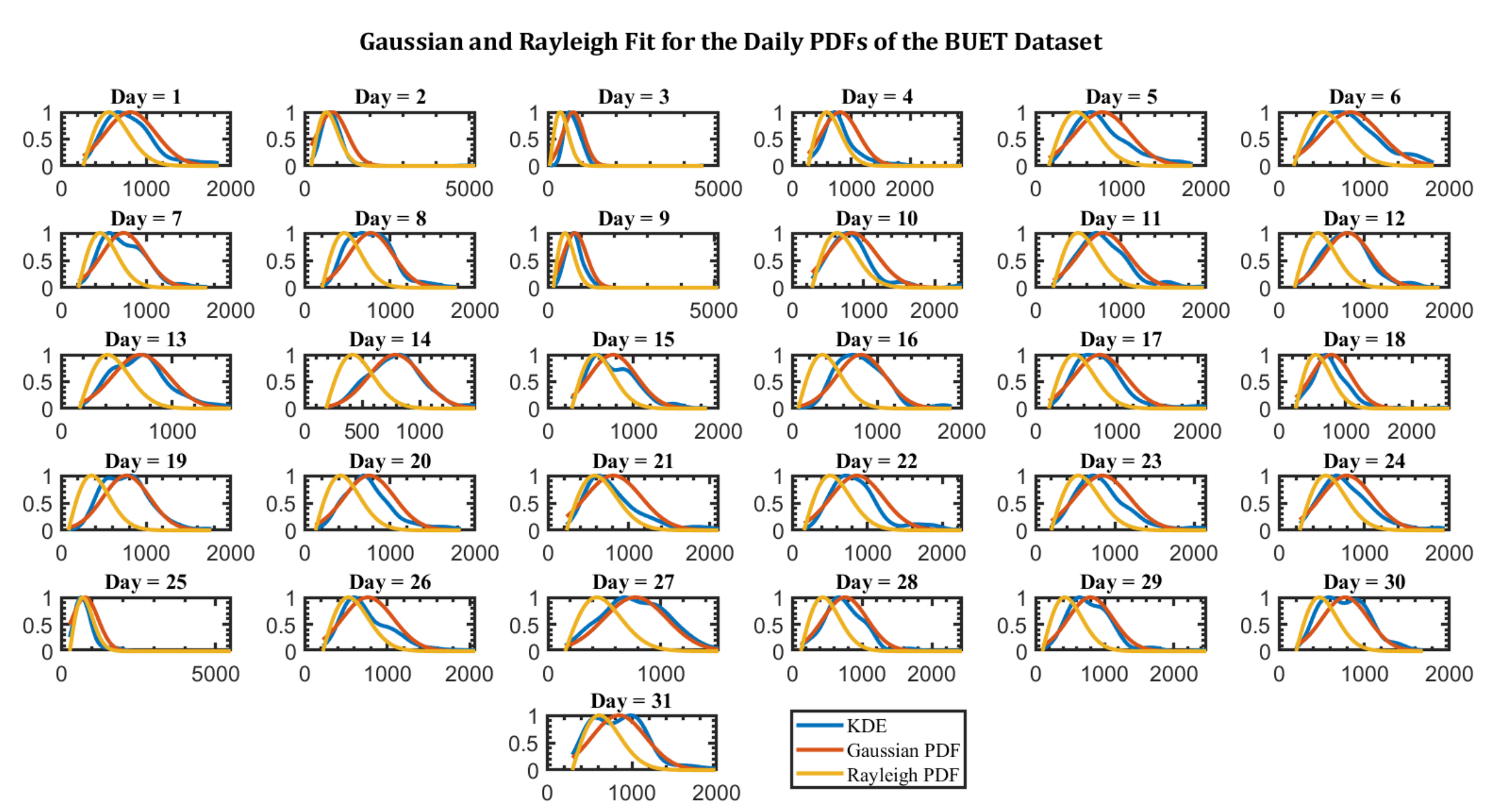}
        \caption{Daily}
        \label{BUET_daily_dist}
    \end{subfigure}
    \hfill
    \begin{subfigure}[t]{0.32\textwidth}
        \centering
        \includegraphics[width=1\textwidth]{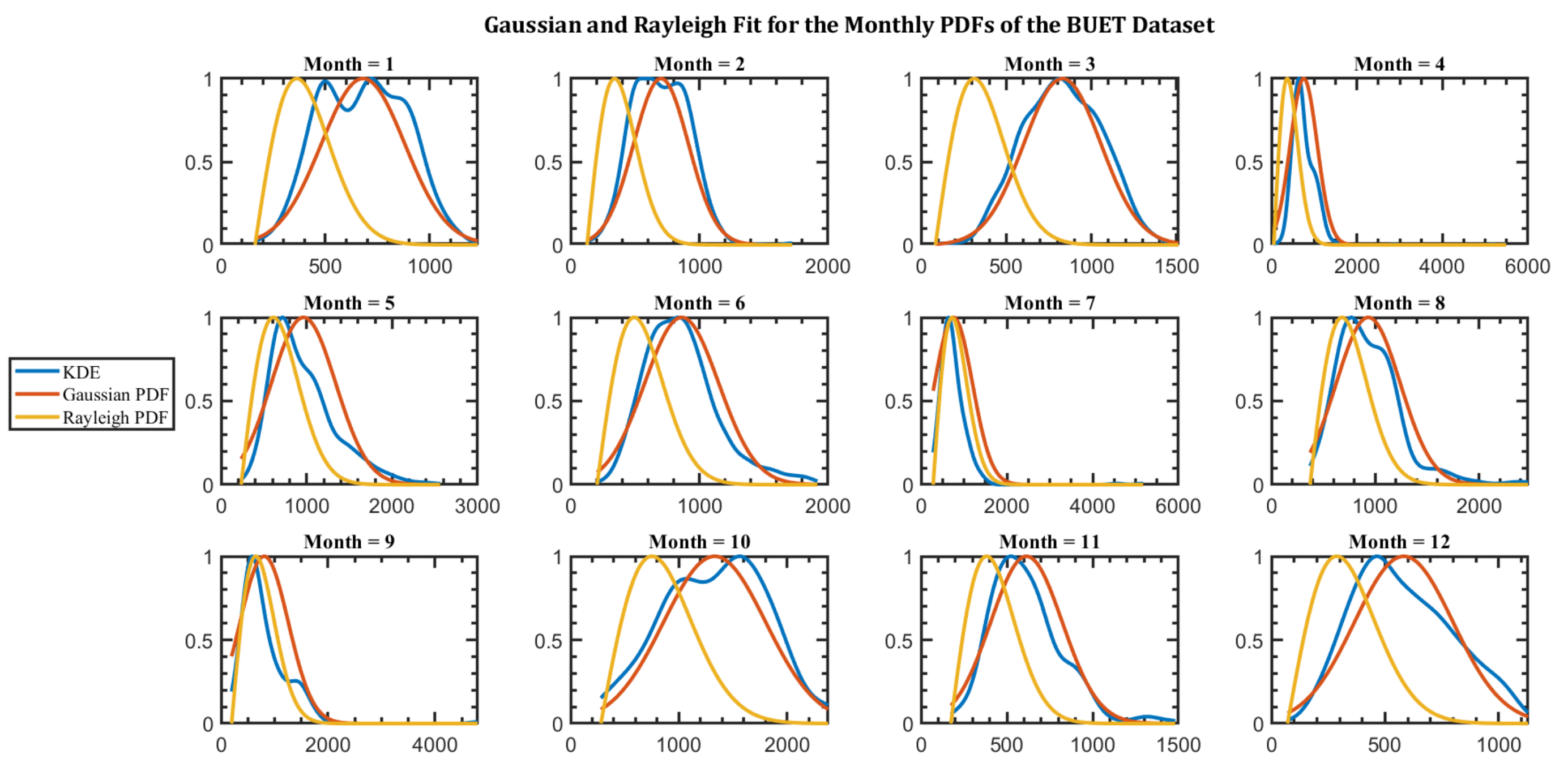}
        \caption{Monthly}
        \label{BUET_monthly_dist}
    \end{subfigure}
    \caption{Gaussian and Rayleigh fit for load probability densities}
    \label{Fit_Load_Probability_Densities}
\end{figure}

Fig.~\ref{BUET_hourly_dist}, ~\ref{BUET_daily_dist}, and ~\ref{BUET_monthly_dist} present a comparative fit of the hourly, daily, and monthly KDEs with Gaussian and Rayleigh DFs. The results indicate that the Gaussian DF provides a closer approximation to the KDEs than the Rayleigh DF across all time scales. To further assess the goodness-of-fit, a Chi-Squared test \cite{greenwood1996guide} was conducted, with the null hypothesis assuming that the data follows a Gaussian distribution. Although the obtained $p$-value was below the significance threshold, under the Central Limit Theorem \cite{dudley2014uniform}, the dataset distribution was considered to be Gaussian.

\subsection{Data Visualization}
\label{subsec6}

\begin{figure}[t]
    \centering
    \begin{minipage}{0.49\textwidth}
    \centering
    \includegraphics[width=1\textwidth]{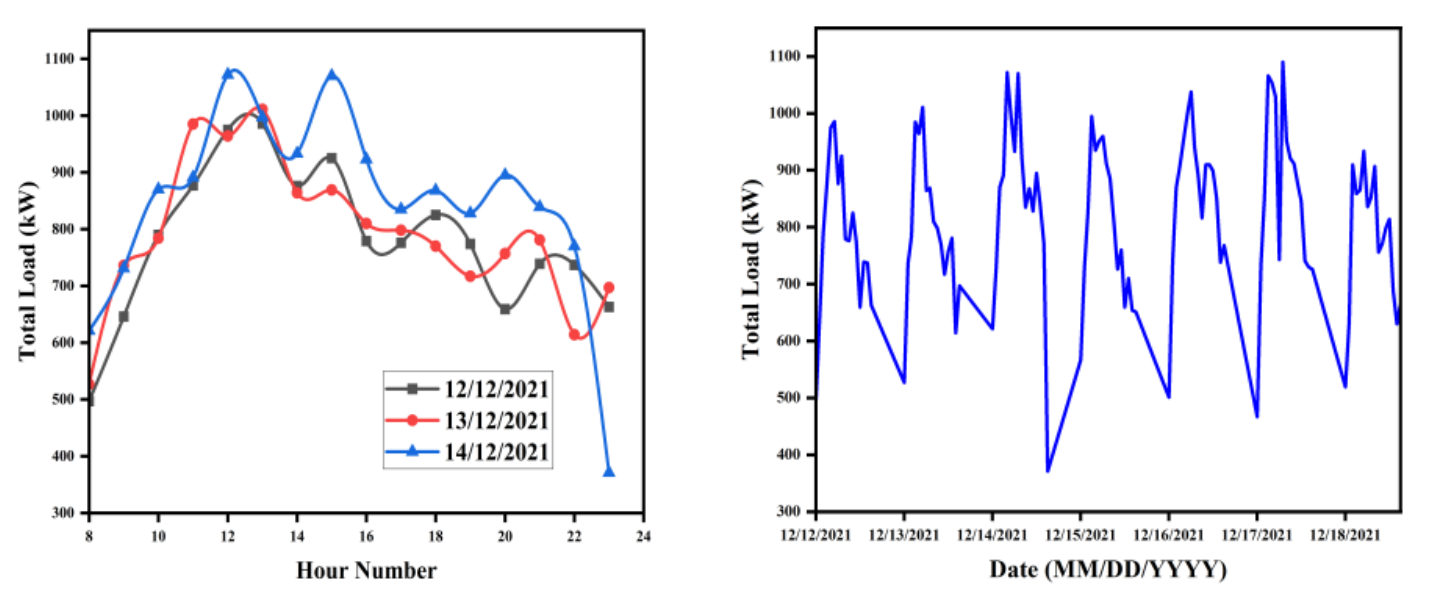}
    \caption{Hourly and weekly variation of the interpolated data for one day}
    \label{BUET_h_variation_1_day_week}
    \end{minipage}
    \hfill
    \begin{minipage}{0.49\textwidth}
    \centering
    \includegraphics[width=1\textwidth]{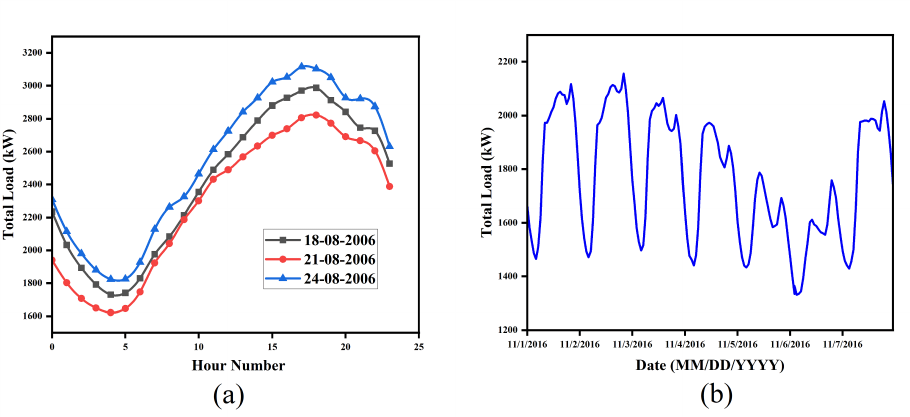}
    \caption{Hourly and weekly variation of the Dayton dataset for one week}
    \label{Dayton_variation}
    \end{minipage}   
\end{figure}

The hourly and weekly variations of the load data (post-interpolation) are illustrated in Fig.~\ref{BUET_h_variation_1_day_week}. The periodic nature of the dataset is evident, with load demand generally increasing during midday and decreasing at night. However, additional factors may influence this trend, as observed in the load demand pattern on March 12, 2021. Fig.~\ref{Dayton_variation} presents the daily and weekly load variations of the Dayton dataset. A distinct pattern emerges where the hourly load demand reaches a minimum during nighttime, when consumption is at its lowest, followed by a peak during the day. Furthermore, the daily demand exhibits a well-defined periodicity, similar to that observed in the BUET power plant dataset.

\subsection{Performance Indices for Prediction Model}
\label{subsec7}

\begin{figure}[t]
\includegraphics[width=1\textwidth]{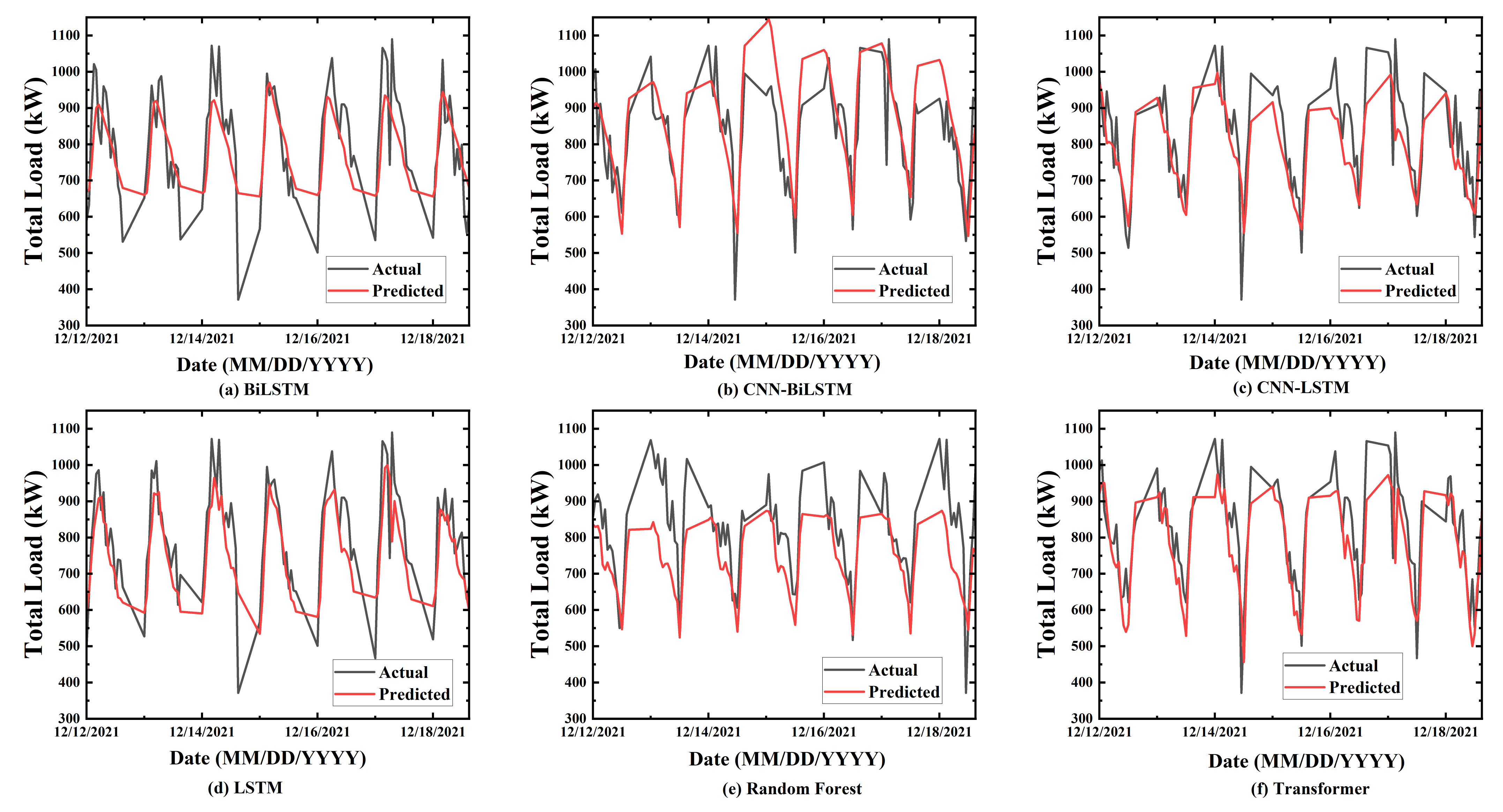}
\caption{Predicted and actual data for different prediction schemes}
\label{predicted_v_actual_BUET}
\end{figure}

Mean Absolute Error (MAE) and Mean Absolute Percentage Error (MAPE) were used as the indices~\cite{de2016mean} to evaluate the performance of all the models under consideration. The MAE and MAPE are defined as follows:
\begin{align}
    \text{MAE} = \frac{1}{n} \sum_{i=1}^{n} \left| y_i - \hat{y}_i \right|, \\
    \text{MAPE} = \frac{1}{n} \sum_{i=1}^{n} \left| \frac{y_i - \hat{y}_i}{y_i} \right|\times100\%
\end{align}
where, $y_i$ is the actual value, $\hat{y_i}$ is the predicted value and $n$ is the number of data points. Fig.~\ref{predicted_v_actual_BUET} compares various prediction schemes by plotting on the same axes the actual and predicted loads.

\subsection{Performance of RNN-based Models}
\label{subsec8}

Purely RNN-based models demonstrated the best overall performance on the dataset. The LSTM and BiLSTM models achieved MAPE of 10.67\% and 13.13\%, respectively, while the Transformer model attained a MAPE of 11.02\%. Incorporating a shallow one-dimensional CNN for feature extraction prior to the LSTM did not yield significant improvements in performance. Similarly, the CNN-BiLSTM model did not exhibit superior results. These findings indicate that purely RNN-based architectures are more effective, suggesting that future load values are predominantly influenced by sequential dependencies within a given window of past loads and features. The LSTM model, in particular, accurately captures the overall trends and seasonal variations in the load data. However, while it effectively models major patterns, it struggles to precisely predict the residual variations in load demand.

\subsection{Performance of Non-RNN Models}
\label{subsec9}

Among the non-RNN neural network models, DLinear exhibited the best performance, achieving a MAPE of 13.13\%, whereas Prophet obtained a MAPE of 17.73\%. Notably, the performance of DLinear was comparable to that of the Transformer and LSTM models, despite being a linear model \cite{zeng2023transformers}.

\subsection{Performance of Classical ML Models}
\label{subsec10}

\begin{figure}[t]
    \centering
    \begin{minipage}{0.49\textwidth}
    \centering
    \includegraphics[width=1\textwidth]{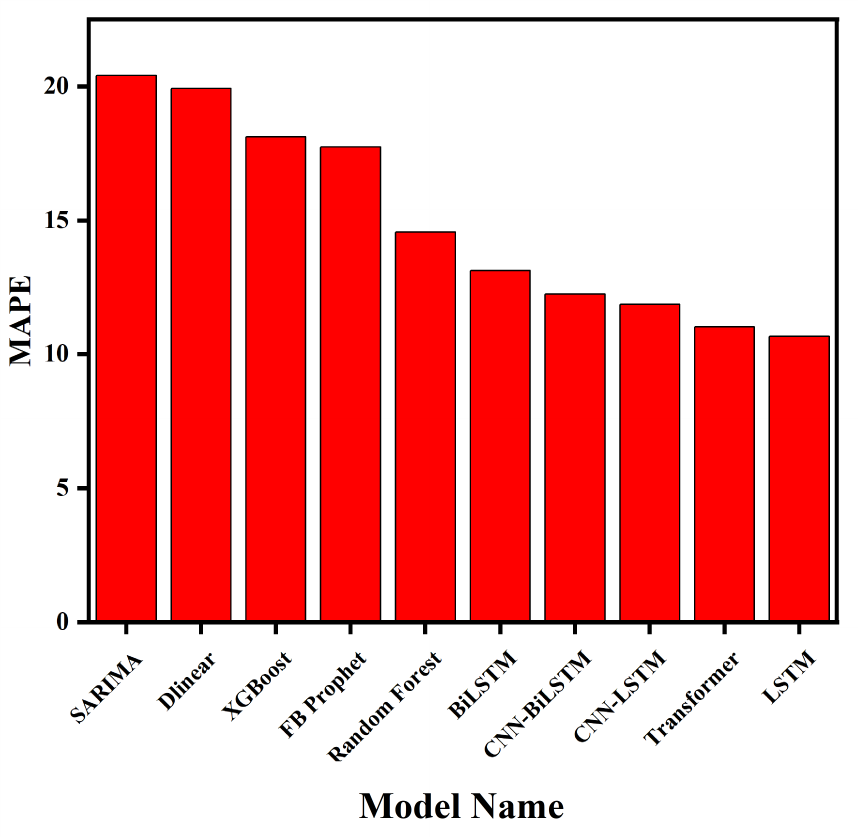}
    \caption{MAPE for various prediction schemes}
    \label{comparison}
    \end{minipage}
    \hfill
    \begin{minipage}{0.49\textwidth}
    \centering
    \includegraphics[width=1\textwidth]{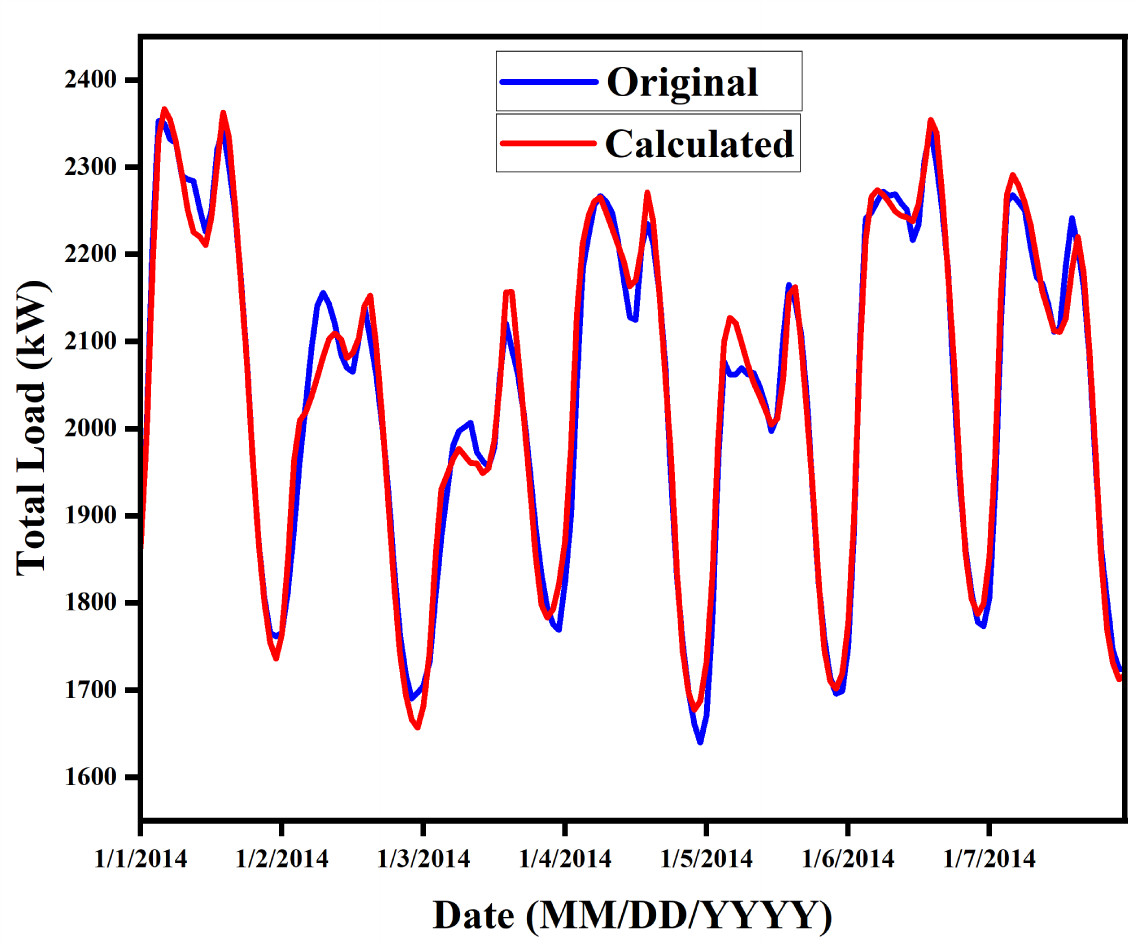}
    \caption{Predicted and Actual Load Values for the Dayton dataset using LSTM prediction scheme}
    \label{LSTM_Dayton}
    \end{minipage}   
\end{figure}

SARIMA, Random Forest, and XGBoost are the classical ML models that we consider in this work. These models exhibited relatively lower performance on the dataset, with Random Forest achieving the best results among them, obtaining a MAPE of 14.56\%. In comparison, XGBoost and SARIMA yielded MAPE values of 18.11\% and 20.40\%, respectively. These findings indicate that traditional machine learning models struggled to effectively capture the variations present in the augmented load dataset. Fig. \ref{comparison} shows a comparison of MAPE of different load forecasting models trained on this sparse dataset

\subsection{Performance Comparison of LSTM-based Models Multiple Datasets}
\label{subsec11}

Since LSTM demonstrated the best MAPE, this model was trained on the Dayton dataset and the results are compared with those obtained from the augmented dataset. Fig.~\ref{LSTM_Dayton} illustrates the predicted and actual load values for a week at the Dayton power plant using the LSTM model. The results show that the predictions closely align with the actual data, capturing both trend, seasonal, and residual variations, thereby validating our approach. The LSTM model achieved a MAPE of 1.55\% and a Mean Absolute Error (MAE) of 31.06\% on this dataset. A comparison with Fig.~\ref{comparison} (d) further confirms that the LSTM model consistently outperforms other models across both datasets.

\begin{figure}[t]
\includegraphics[width=1\textwidth]{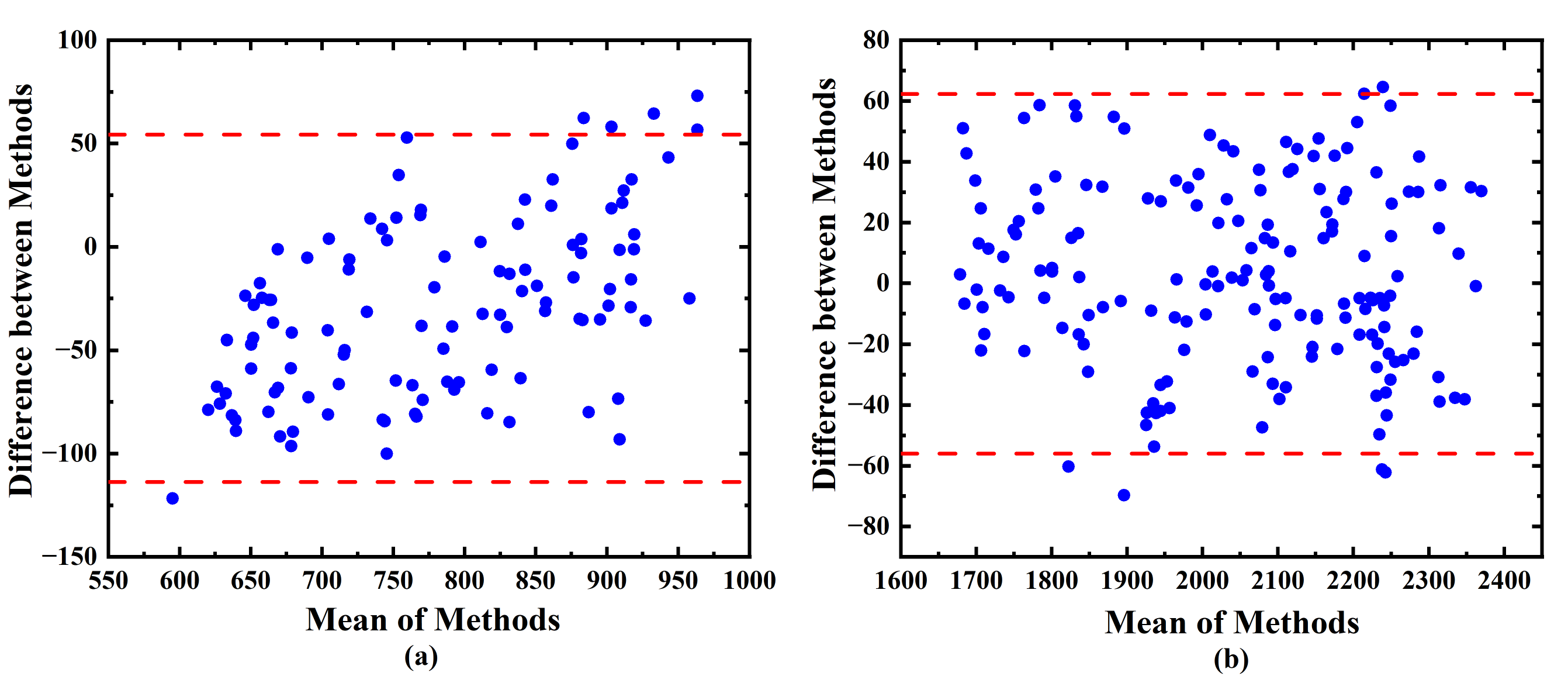}
\caption{Bland-Altman plots between LSTM and BiLSTM models for (a) interpolated BUET dataset (b) Dayton dataset}
\label{Bland-Altman}
\end{figure}

To assess the variation in predictions across different models, Bland-Altman Plots (BAP) \cite{giavarina2015understanding} were computed to compare the LSTM and BiLSTM models on both datasets, as shown in Fig.~\ref{Bland-Altman}. The BAP illustrates the difference ($y$) between the predictions of two models as a function of the mean ($x$) of their corresponding predictions. When the BAP is centered around the $y = 0$ horizontal line, it indicates that both models produce nearly identical predictions across all predicted values. Conversely, a more dispersed BAP suggests greater disagreement between models. Additionally, if the spread increases towards higher $x$ values, it implies that the models diverge more significantly at higher predicted values.

For the BUET dataset, the BAP reveals that LSTM tends to predict higher values than BiLSTM ($y_{\text{LSTM}} > y_{\text{BiLSTM}}$) for larger load values, whereas for lower to medium load predictions, LSTM predicts lower values than BiLSTM ($y_{\text{LSTM}} < y_{\text{BiLSTM}}$). In contrast, for the Dayton dataset, the BAP exhibits an almost equal distribution of data points on both sides of the zero line across all load prediction values. This suggests that both models perform similarly on the Dayton dataset. These findings indicate that predictions obtained from augmenting a highly sparse dataset are largely model-specific.

\section{Conclusion}
\label{sec5}

Existing research on sparse non-stationary datasets, particularly in the context of load forecasting, is very limited due to the inherent challenges associated with such datasets. In this work, we addressed this research gap by utilizing a novel interpolation approach for estimating the missing data by assuming that the data generation is a WSS process. We utilized several classical signal processing and machine learning approaches and some state of the art neural network based-models for load forecasting using the interpolated data, and showed that the proposed data interpolation technique does provide excellent performance. Throughout this study, all methods applied to the dataset were confined to the time domain, with no transformations into alternative domains. Advanced compressed sensing techniques, such as Fourier or wavelet transforms, may provide valuable insights; however, the significant sparsity in the time domain could introduce additional challenges. Future research using this dataset can focus on identifying seasonality trends within the sparse dataset and leveraging them to enhance interpolation accuracy. Moreover, one may also explore modifications to the model architectures or the adoption of novel models to improve predictive performance. Additionally, alternative interpolation and transformation techniques, including filtering methods such as Fourier or wavelet transforms, could be investigated to achieve more robust interpolation and prediction outcomes.

\section{Data Availability}
\label{sec6}

The hourly load data for the years 2015, 2016, 2021 and 2022 for the BUET Power Plant and the codes for this study may be availed upon reasonable request from the authors.

\section{Acknowledgment}
\label{sec7}

The authors acknowledge the support of the BUET power plant located in West Palashi, Dhaka, Bangladesh; Dr. Nasim Ahmed Dewan, Shafayeth Jamil and Akif Hamid from the Department of Electrical and Electronic Engineering, BUET, for this research. The weather data was obtained from the National Aeronautics and Space Administration (NASA) Langley Research Center (LaRC) Prediction of Worldwide Energy Resource (POWER) Project funded through the NASA Earth Science/Applied Science Program. The PJM Hourly Load Dataset is licensed publicly under the Creative Commons license CC0.




\begin{thebibliography}{10}
\expandafter\ifx\csname url\endcsname\relax
  \def\url#1{\texttt{#1}}\fi
\expandafter\ifx\csname urlprefix\endcsname\relax\def\urlprefix{URL }\fi
\expandafter\ifx\csname href\endcsname\relax
  \def\href#1#2{#2} \def\path#1{#1}\fi

\bibitem{palanivinayagam2023missing}
A.~Palanivinayagam, R.~Damaševičius, Effective handling of missing values in datasets for classification using machine learning methods, Information 14~(2) (2023).
\newblock \href {https://doi.org/10.3390/info14020092} {\path{doi:10.3390/info14020092}}.

\bibitem{gorban2018blessing}
A.~N. Gorban, I.~Y. Tyukin, Blessing of dimensionality: mathematical foundations of the statistical physics of data, Philosophical Transactions of the Royal Society A: Mathematical, Physical and Engineering Sciences 376~(2118) (2018) 20170237.
\newblock \href {https://doi.org/10.1098/rsta.2017.0237} {\path{doi:10.1098/rsta.2017.0237}}.

\bibitem{fefferman2016testing}
C.~Fefferman, S.~Mitter, H.~Narayanan, Testing the manifold hypothesis, Journal of the American Mathematical Society 29~(4) (2016) 983--1049.
\newblock \href {https://doi.org/10.1090/jams/852} {\path{doi:10.1090/jams/852}}.

\bibitem{donoho2006compressed}
D.~L. Donoho, Compressed sensing, IEEE Transactions on information theory 52~(4) (2006) 1289--1306.
\newblock \href {https://doi.org/10.1109/TIT.2006.871582} {\path{doi:10.1109/TIT.2006.871582}}.

\bibitem{candes2006robust}
E.~J. Cand{\`e}s, J.~Romberg, T.~Tao, Robust uncertainty principles: Exact signal reconstruction from highly incomplete frequency information, IEEE Transactions on information theory 52~(2) (2006) 489--509.
\newblock \href {https://doi.org/10.1109/TIT.2005.862083} {\path{doi:10.1109/TIT.2005.862083}}.

\bibitem{bunn1982short}
D.~W. Bunn, Short-term forecasting: A review of procedures in the electricity supply industry, The Journal of the Operational Research Society 33~(6) (1982) 533--545.
\newblock \href {https://doi.org/10.2307/2581037} {\path{doi:10.2307/2581037}}.

\bibitem{xu2025deep}
C.~Xu, Q.~Li, T.~Xiao, Y.~Zhang, W.~Zhou, H.~Liu, Deep learning-based post-disaster energy management and faster network reconfiguration method for improvement of restoration time, Electric Power Systems Research 238 (2025) 111081.

\bibitem{haida2002regression}
T.~Haida, S.~Muto, Regression based peak load forecasting using a transformation technique (1994).
\newblock \href {https://doi.org/10.1109/59.331433} {\path{doi:10.1109/59.331433}}.

\bibitem{dehbozorgi2025deep}
M.~R. Dehbozorgi, M.~Rastegar, et~al., A deep learning deviation-based scheme to defend against false data injection attacks in power distribution systems, Electric Power Systems Research 238 (2025) 111076.

\bibitem{he2025short}
X.~He, W.~Zhao, Z.~Gao, L.~Zhang, Q.~Zhang, X.~Li, Short-term load forecasting by gru neural network and ddpg algorithm for adaptive optimization of hyperparameters, Electric Power Systems Research 238 (2025) 111119.

\bibitem{Singh2012LoadFT}
A.~K. Singh, Ibraheem, S.~Khatoon, M.~Muazzam, D.~K. Chaturvedi, Load forecasting techniques and methodologies: A review, 2012 2nd International Conference on Power, Control and Embedded Systems (2012) 1--10\href {https://doi.org/10.1109/ICPCES.2012.6508132} {\path{doi:10.1109/ICPCES.2012.6508132}}.

\bibitem{Papalexopoulos1989regression}
A.~Papalexopoulos, T.~Hesterberg, A regression-based approach to short-term system load forecasting, in: Conference Papers Power Industry Computer Application Conference, 1989, pp. 414--423.
\newblock \href {https://doi.org/10.1109/PICA.1989.39025} {\path{doi:10.1109/PICA.1989.39025}}.

\bibitem{Christiaanse1971Short}
W.~R. Christiaanse, Short-term load forecasting using general exponential smoothing, IEEE Transactions on Power Apparatus and Systems PAS-90~(2) (1971) 900--911.
\newblock \href {https://doi.org/10.1109/TPAS.1971.293123} {\path{doi:10.1109/TPAS.1971.293123}}.

\bibitem{Timothy1995neural}
T.~Masters, Neural, Novel and Hybrid Algorithms for Time Series Prediction, 1st Edition, John Wiley \& Sons, Inc., USA, 1995.
\newblock \href {https://doi.org/10.5555/526065} {\path{doi:10.5555/526065}}.

\bibitem{sadaei2019short}
H.~J. Sadaei, P.~C. {de Lima e Silva}, F.~G. Guimarães, M.~H. Lee, Short-term load forecasting by using a combined method of convolutional neural networks and fuzzy time series, Energy 175 (2019) 365--377.
\newblock \href {https://doi.org/10.1016/j.energy.2019.03.081} {\path{doi:10.1016/j.energy.2019.03.081}}.

\bibitem{priyadarsini2025cnn}
M.~Priyadarsini, N.~Sonekar, A cnn-based approach for anomaly detection in smart grid systems, Electric Power Systems Research 238 (2025) 111077.

\bibitem{Aseeri2023effective}
A.~O. Aseeri, Effective rnn-based forecasting methodology design for improving short-term power load forecasts: Application to large-scale power-grid time series, Journal of Computational Science 68 (2023) 101984.
\newblock \href {https://doi.org/10.1016/j.jocs.2023.101984} {\path{doi:10.1016/j.jocs.2023.101984}}.

\bibitem{Li2022research}
C.~Li, Q.~Guo, L.~Shao, J.~Li, H.~Wu, Research on short-term load forecasting based on optimized gru neural network, Electronics 11~(22) (2022).
\newblock \href {https://doi.org/10.3390/electronics11223834} {\path{doi:10.3390/electronics11223834}}.

\bibitem{Pengdan2024novel}
P.~Fan, D.~Wang, W.~Wang, X.~Zhang, Y.~Sun, A novel multi-energy load forecasting method based on building flexibility feature recognition technology and multi-task learning model integrating lstm, Energy 308 (2024) 132976.
\newblock \href {https://doi.org/10.1016/j.energy.2024.132976} {\path{doi:10.1016/j.energy.2024.132976}}.

\bibitem{tulensalo2020lstm}
J.~Tulensalo, J.~Sepp{\"a}nen, A.~Ilin, An lstm model for power grid loss prediction, Electric Power Systems Research 189 (2020) 106823.

\bibitem{rosseel2025physics}
A.~Rosseel, B.~B. Zad, F.~Vall{\'e}e, Z.~De~Gr{\`e}ve, Physics-informed machine learning for forecasting power exchanges at the interface between transmission and distribution systems, Electric Power Systems Research 238 (2025) 111097.

\bibitem{huang2021missing}
G.~Huang, Missing data filling method based on linear interpolation and lightgbm, in: Journal of Physics: Conference Series, Vol. 1754, IOP Publishing, 2021, p. 012187.
\newblock \href {https://doi.org/10.1088/1742-6596/1754/1/012187} {\path{doi:10.1088/1742-6596/1754/1/012187}}.

\bibitem{noor2014filling}
M.~Noor, A.~Yahaya, N.~A. Ramli, A.~M. Al~Bakri, Filling missing data using interpolation methods: Study on the effect of fitting distribution, Key Engineering Materials 594 (2014) 889--895.
\newblock \href {https://doi.org/10.4028/www.scientific.net/KEM.594-595.889} {\path{doi:10.4028/www.scientific.net/KEM.594-595.889}}.

\bibitem{thompson1952construction}
W.~R. Thompson, C.~S. Weil, On the construction of tables for moving-average interpolation, Biometrics 8~(1) (1952) 51--54.
\newblock \href {https://doi.org/10.2307/3001525} {\path{doi:10.2307/3001525}}.

\bibitem{gomez1994estimation}
V.~G{\'o}mez, A.~Maravall, Estimation, prediction, and interpolation for nonstationary series with the kalman filter, Journal of the American Statistical Association 89~(426) (1994) 611--624.
\newblock \href {https://doi.org/10.2307/2290864} {\path{doi:10.2307/2290864}}.

\bibitem{kai2008nurbs}
Z.~Kai, W.~Guanjun, J.~Houzhong, T.~Zhongyi, Nurbs interpolation based on exponential smoothing forecasting, The International Journal of Advanced Manufacturing Technology 39 (2008) 1190--1196.
\newblock \href {https://doi.org/10.1007/s00170-007-1297-0} {\path{doi:10.1007/s00170-007-1297-0}}.

\bibitem{wahba1981spline}
G.~Wahba, Spline interpolation and smoothing on the sphere, SIAM Journal on Scientific and Statistical Computing 2~(1) (1981) 5--16.
\newblock \href {https://doi.org/10.1137/0902002} {\path{doi:10.1137/0902002}}.

\bibitem{rabbath2019comparison}
C.~Rabbath, D.~Corriveau, A comparison of piecewise cubic hermite interpolating polynomials, cubic splines and piecewise linear functions for the approximation of projectile aerodynamics, Defence Technology 15~(5) (2019) 741--757.
\newblock \href {https://doi.org/10.1016/j.dt.2019.07.016} {\path{doi:10.1016/j.dt.2019.07.016}}.

\bibitem{tjostheim1975some}
D.~Tjostheim, J.~Thomas, Some properties and examples of random processes that are almost wide sense stationary, IEEE Transactions on Information Theory 21~(3) (1975) 257--262.
\newblock \href {https://doi.org/10.1109/TIT.1975.1055385} {\path{doi:10.1109/TIT.1975.1055385}}.

\bibitem{priestley1967power}
M.~Priestley, Power spectral analysis of non-stationary random processes, Journal of Sound and Vibration 6~(1) (1967) 86--97.
\newblock \href {https://doi.org/10.1016/0022-460X(67)90160-5} {\path{doi:10.1016/0022-460X(67)90160-5}}.

\bibitem{edition2002probability}
F.~Edition, A.~Papoulis, S.~U. Pillai, Probability, random variables, and stochastic processes, McGraw-Hill Europe: New York, NY, USA, 2002.
\newblock \href {https://doi.org/10.1109/TASSP.1985.1164715} {\path{doi:10.1109/TASSP.1985.1164715}}.

\bibitem{mills2011dealing}
T.~C. Mills, T.~C. Mills, Dealing with nonstationarity: Detrending, smoothing and differencing, The Foundations of Modern Time Series Analysis (2011) 261--288\href {https://doi.org/10.1057/9780230305021_10} {\path{doi:10.1057/9780230305021_10}}.

\bibitem{dombi2020new}
J.~Dombi, A.~Hussain, A new approach to fuzzy control using the distending function, Journal of Process Control 86 (2020) 16--29.
\newblock \href {https://doi.org/10.1016/j.jprocont.2019.12.005} {\path{doi:10.1016/j.jprocont.2019.12.005}}.

\bibitem{poulinakis2023machine}
K.~Poulinakis, D.~Drikakis, I.~W. Kokkinakis, S.~M. Spottswood, Machine-learning methods on noisy and sparse data, Mathematics 11~(1) (2023) 236.
\newblock \href {https://doi.org/10.3390/math11010236} {\path{doi:10.3390/math11010236}}.

\bibitem{huang2015bidirectional}
Z.~Huang, W.~Xu, K.~Yu, Bidirectional lstm-crf models for sequence tagging, arXiv preprint arXiv:1508.01991 (2015).
\newblock \href {https://doi.org/10.48550/arXiv.1508.01991} {\path{doi:10.48550/arXiv.1508.01991}}.

\bibitem{lu2020cnn}
W.~Lu, J.~Li, Y.~Li, A.~Sun, J.~Wang, A cnn-lstm-based model to forecast stock prices, Complexity 2020~(1) (2020) 6622927.
\newblock \href {https://doi.org/10.1155/2020/6622927} {\path{doi:10.1155/2020/6622927}}.

\bibitem{lu2021cnn}
W.~Lu, J.~Li, J.~Wang, L.~Qin, A cnn-bilstm-am method for stock price prediction, Neural Computing and Applications 33~(10) (2021) 4741--4753.
\newblock \href {https://doi.org/10.1007/s00521-020-05532-z} {\path{doi:10.1007/s00521-020-05532-z}}.

\bibitem{vaswani2017attention}
A.~Vaswani, Attention is all you need, Advances in Neural Information Processing Systems (2017).

\bibitem{jha2021time}
B.~K. Jha, S.~Pande, Time series forecasting model for supermarket sales using fb-prophet, in: 2021 5th International Conference on Computing Methodologies and Communication (ICCMC), IEEE, 2021, pp. 547--554.
\newblock \href {https://doi.org/10.1109/ICCMC51019.2021.9418033} {\path{doi:10.1109/ICCMC51019.2021.9418033}}.

\bibitem{zeng2023transformers}
A.~Zeng, M.~Chen, L.~Zhang, Q.~Xu, Are transformers effective for time series forecasting?, in: Proceedings of the AAAI conference on artificial intelligence, Vol.~37, 2023, pp. 11121--11128.
\newblock \href {https://doi.org/10.1609/aaai.v37i9.26317} {\path{doi:10.1609/aaai.v37i9.26317}}.

\bibitem{Chen2018time}
P.~Chen, A.~Niu, D.~Liu, W.~Jiang, B.~Ma, Time series forecasting of temperatures using sarima: An example from nanjing, IOP Conference Series: Materials Science and Engineering 394 (2018) 052024.
\newblock \href {https://doi.org/10.1088/1757-899X/394/5/052024} {\path{doi:10.1088/1757-899X/394/5/052024}}.

\bibitem{chen2016xgboost}
T.~Chen, C.~Guestrin, Xgboost: A scalable tree boosting system, in: Proceedings of the 22nd acm sigkdd international conference on knowledge discovery and data mining, 2016, pp. 785--794.
\newblock \href {https://doi.org/10.1145/2939672.2939785} {\path{doi:10.1145/2939672.2939785}}.

\bibitem{rigatti2017random}
S.~J. Rigatti, Random forest, Journal of Insurance Medicine 47~(1) (2017) 31--39.
\newblock \href {https://doi.org/10.17849/insm-47-01-31-39.1} {\path{doi:10.17849/insm-47-01-31-39.1}}.

\bibitem{nasapower}
N.~Aeronautics, S.~A. N. L. R. C. L.~P. of~Worldwide Energy Resource~(POWER), Data access viewer (dav), accessed: Feb 10, 2023 (2023).

\bibitem{pjmdataset}
R.~Molla, Pjm hourly energy consumption data, \url{https://www.kaggle.com/datasets/robikscube/hourly-energy-consumption/data} (2018).

\bibitem{priyambudi2024algorithm}
Z.~S. Priyambudi, Y.~S. Nugroho, Which algorithm is better? an implementation of normalization to predict student performance, in: AIP Conference Proceedings, Vol. 2926, AIP Publishing, 2024.

\bibitem{bence1995analysis}
J.~R. Bence, Analysis of short time series: correcting for autocorrelation, Ecology 76~(2) (1995) 628--639.
\newblock \href {https://doi.org/10.2307/1941218} {\path{doi:10.2307/1941218}}.

\bibitem{xu2019temporal}
C.~Xu, M.~Nayyeri, F.~Alkhoury, H.~S. Yazdi, J.~Lehmann, Temporal knowledge graph embedding model based on additive time series decomposition, arXiv preprint arXiv:1911.07893 (2019).
\newblock \href {https://doi.org/10.48550/arXiv.1911.07893} {\path{doi:10.48550/arXiv.1911.07893}}.

\bibitem{dickey1979distribution}
D.~A. Dickey, W.~A. Fuller, Distribution of the estimators for autoregressive time series with a unit root, Journal of the American statistical association 74~(366a) (1979) 427--431.
\newblock \href {https://doi.org/10.2307/2286348} {\path{doi:10.2307/2286348}}.

\bibitem{kwiatkowski1992testing}
D.~Kwiatkowski, P.~C. Phillips, P.~Schmidt, Y.~Shin, Testing the null hypothesis of stationarity against the alternative of a unit root: How sure are we that economic time series have a unit root?, Journal of econometrics 54~(1-3) (1992) 159--178.

\bibitem{box2015time}
G.~E. Box, G.~M. Jenkins, G.~C. Reinsel, G.~M. Ljung, Time series analysis: forecasting and control, John Wiley \& Sons, 2015.
\newblock \href {https://doi.org/10.1111/jtsa.12194} {\path{doi:10.1111/jtsa.12194}}.

\bibitem{Ng01031995}
S.~Ng, P.~Perron, Unit root tests in arma models with data-dependent methods for the selection of the truncation lag, Journal of the American Statistical Association 90~(429) (1995) 268--281.
\newblock \href {https://doi.org/10.1080/01621459.1995.10476510} {\path{doi:10.1080/01621459.1995.10476510}}.

\bibitem{dubnicka2009kernel}
S.~R. Dubnicka, Kernel density estimation with missing data and auxiliary variables, Australian \& New Zealand Journal of Statistics 51~(3) (2009) 247--270.
\newblock \href {https://doi.org/10.1111/j.1467-842X.2009.00541.x} {\path{doi:10.1111/j.1467-842X.2009.00541.x}}.

\bibitem{greenwood1996guide}
P.~E. Greenwood, M.~S. Nikulin, A guide to chi-squared testing, Vol. 280, John Wiley \& Sons, 1996.
\newblock \href {https://doi.org/10.1080/00224065.1997.11979805} {\path{doi:10.1080/00224065.1997.11979805}}.

\bibitem{dudley2014uniform}
R.~M. Dudley, Uniform central limit theorems, Vol. 142, Cambridge university press, 2014.
\newblock \href {https://doi.org/10.1017/CBO9780511665622} {\path{doi:10.1017/CBO9780511665622}}.

\bibitem{de2016mean}
A.~De~Myttenaere, B.~Golden, B.~Le~Grand, F.~Rossi, Mean absolute percentage error for regression models, Neurocomputing 192 (2016) 38--48.

\bibitem{giavarina2015understanding}
D.~Giavarina, Understanding bland altman analysis, Biochemia medica 25~(2) (2015) 141--151.
\newblock \href {https://doi.org/10.11613/BM.2015.015} {\path{doi:10.11613/BM.2015.015}}.

\end{thebibliography}


\begin{thebibliography}{10}
\expandafter\ifx\csname url\endcsname\relax
  \def\url#1{\texttt{#1}}\fi
\expandafter\ifx\csname urlprefix\endcsname\relax\def\urlprefix{URL }\fi
\expandafter\ifx\csname href\endcsname\relax
  \def\href#1#2{#2} \def\path#1{#1}\fi

\bibitem{taylor2018forecasting}
S.~J. Taylor, B.~Letham, Forecasting at scale, The American Statistician 72~(1) (2018) 37--45.

\bibitem{hochreiter1997long}
S.~Hochreiter, J.~Schmidhuber, Long short-term memory, Neural computation 9~(8) (1997) 1735--1780.

\bibitem{graves2005framewise}
A.~Graves, J.~Schmidhuber, \href{https://www.sciencedirect.com/science/article/pii/S0893608005001206}{Framewise phoneme classification with bidirectional lstm and other neural network architectures}, Neural Networks 18~(5) (2005) 602--610, iJCNN 2005.
\newblock \href {https://doi.org/https://doi.org/10.1016/j.neunet.2005.06.042} {\path{doi:https://doi.org/10.1016/j.neunet.2005.06.042}}.
\newline\urlprefix\url{https://www.sciencedirect.com/science/article/pii/S0893608005001206}

\bibitem{huang2015bidirectional}
Z.~Huang, W.~Xu, K.~Yu, Bidirectional lstm-crf models for sequence tagging, arXiv preprint arXiv:1508.01991 (2015).
\newblock \href {https://doi.org/10.48550/arXiv.1508.01991} {\path{doi:10.48550/arXiv.1508.01991}}.

\bibitem{graves2005bidirectional}
A.~Graves, S.~Fern{\'a}ndez, J.~Schmidhuber, Bidirectional lstm networks for improved phoneme classification and recognition, in: International conference on artificial neural networks, Springer, 2005, pp. 799--804.

\bibitem{shi2015convolutional}
X.~Shi, Z.~Chen, H.~Wang, D.-Y. Yeung, W.~kin Wong, W.~chun Woo, \href{https://arxiv.org/abs/1506.04214}{Convolutional lstm network: A machine learning approach for precipitation nowcasting} (2015).
\newblock \href {http://arxiv.org/abs/1506.04214} {\path{arXiv:1506.04214}}.
\newline\urlprefix\url{https://arxiv.org/abs/1506.04214}

\bibitem{lu2020cnn}
W.~Lu, J.~Li, Y.~Li, A.~Sun, J.~Wang, A cnn-lstm-based model to forecast stock prices, Complexity 2020~(1) (2020) 6622927.
\newblock \href {https://doi.org/10.1155/2020/6622927} {\path{doi:10.1155/2020/6622927}}.

\bibitem{agarap2018deep}
A.~F. Agarap, Deep learning using rectified linear units (relu), arXiv preprint arXiv:1803.08375 (2018).

\bibitem{lu2021cnn}
W.~Lu, J.~Li, J.~Wang, L.~Qin, A cnn-bilstm-am method for stock price prediction, Neural Computing and Applications 33~(10) (2021) 4741--4753.
\newblock \href {https://doi.org/10.1007/s00521-020-05532-z} {\path{doi:10.1007/s00521-020-05532-z}}.

\bibitem{huang2023understanding}
Z.~Huang, M.~Liang, J.~Qin, S.~Zhong, L.~Lin, Understanding self-attention mechanism via dynamical system perspective, in: Proceedings of the IEEE/CVF International Conference on Computer Vision, 2023, pp. 1412--1422.

\bibitem{vaswani2017attention}
A.~Vaswani, Attention is all you need, Advances in Neural Information Processing Systems (2017).

\bibitem{cordonnier2020multi}
J.-B. Cordonnier, A.~Loukas, M.~Jaggi, Multi-head attention: Collaborate instead of concatenate, arXiv preprint arXiv:2006.16362 (2020).

\bibitem{jha2021time}
B.~K. Jha, S.~Pande, Time series forecasting model for supermarket sales using fb-prophet, in: 2021 5th International Conference on Computing Methodologies and Communication (ICCMC), IEEE, 2021, pp. 547--554.
\newblock \href {https://doi.org/10.1109/ICCMC51019.2021.9418033} {\path{doi:10.1109/ICCMC51019.2021.9418033}}.

\bibitem{zeng2023transformers}
A.~Zeng, M.~Chen, L.~Zhang, Q.~Xu, Are transformers effective for time series forecasting?, in: Proceedings of the AAAI conference on artificial intelligence, Vol.~37, 2023, pp. 11121--11128.
\newblock \href {https://doi.org/10.1609/aaai.v37i9.26317} {\path{doi:10.1609/aaai.v37i9.26317}}.

\bibitem{Chen2018time}
P.~Chen, A.~Niu, D.~Liu, W.~Jiang, B.~Ma, Time series forecasting of temperatures using sarima: An example from nanjing, IOP Conference Series: Materials Science and Engineering 394 (2018) 052024.
\newblock \href {https://doi.org/10.1088/1757-899X/394/5/052024} {\path{doi:10.1088/1757-899X/394/5/052024}}.

\bibitem{box2015time}
G.~E. Box, G.~M. Jenkins, G.~C. Reinsel, G.~M. Ljung, Time series analysis: forecasting and control, John Wiley \& Sons, 2015.
\newblock \href {https://doi.org/10.1111/jtsa.12194} {\path{doi:10.1111/jtsa.12194}}.

\bibitem{anderson1976backshift}
O.~Anderson, The backshift operator in time series analysis, International Journal of Mathematical Educational in Science and Technology 7~(2) (1976) 235--241.

\bibitem{nandi2020data}
A.~K. Nandi, Data modeling with polynomial representations and autoregressive time-series representations, and their connections, Ieee Access 8 (2020) 110412--110424.

\bibitem{shao2015effects}
Y.-H. Shao, G.-F. Gu, Z.-Q. Jiang, W.-X. Zhou, Effects of polynomial trends on detrending moving average analysis, Fractals 23~(03) (2015) 1550034.

\bibitem{ahmad2018tree}
M.~W. Ahmad, M.~Mourshed, Y.~Rezgui, Tree-based ensemble methods for predicting pv power generation and their comparison with support vector regression, Energy 164 (2018) 465--474.

\bibitem{chen2016xgboost}
T.~Chen, C.~Guestrin, Xgboost: A scalable tree boosting system, in: Proceedings of the 22nd acm sigkdd international conference on knowledge discovery and data mining, 2016, pp. 785--794.
\newblock \href {https://doi.org/10.1145/2939672.2939785} {\path{doi:10.1145/2939672.2939785}}.

\bibitem{rigatti2017random}
S.~J. Rigatti, Random forest, Journal of Insurance Medicine 47~(1) (2017) 31--39.
\newblock \href {https://doi.org/10.17849/insm-47-01-31-39.1} {\path{doi:10.17849/insm-47-01-31-39.1}}.

\end{thebibliography}

\putbib               
\end{bibunit}

\clearpage
\section*{Supplementary Material}
\phantomsection 
\addcontentsline{toc}{section}{Supplementary Material}

\setcounter{section}{0}
\setcounter{subsection}{0}
\setcounter{figure}{0}
\setcounter{table}{0}
\renewcommand\thesection{S\arabic{section}}
\renewcommand\thefigure{S\arabic{figure}}
\renewcommand\thetable{S\arabic{table}}

\begin{bibunit}

\section{Review of Models}\label{app:background}
In this section, we review the classical and neural network models that we utilized in this work.
\subsection{Long Short-Term Memory (LSTM)}
\label{sec:lstm}

An LSTM network is a type of recurrent neural network (RNN) that is ideal for capturing long-term dependencies in sequential data \cite{taylor2018forecasting}. Traditional RNNs often have problems of vanishing or exploding gradients during training, which LSTMs mitigate by introducing a memory cell capable of preserving information over long sequences. Each LSTM unit processes the input using a set of gating mechanisms that maintain the flow of information. In the following we review the basics of an LSTM network~\cite{hochreiter1997long}.

\begin{itemize}
    \item \textbf{Input Layer:} Inputs to an LSTM network are a time series of length $T$, represented as $\{x_1, x_2, \dots, x_T\}$, where each $x_t \in \mathbb{R}^d$ is a $d$-dimensional feature vector.

    \item \textbf{LSTM Layer:} Each cell at time step $t$ takes $x_t$, the previous hidden state $h_{t-1} \in (-1,1)^h$, and previous cell state $c_{t-1} \in \mathbb{R}^h$ to compute the current output and updated memory. The dimension of the hidden state vector and cell state vector is $h$. The actions of the gates in an LSTM are input, forget and output, along with other actions, are described below:

    \begin{itemize}
        \item \textbf{Cell input activation:}
        \[
        z_t = g(W_z x_t + R_z h_{t-1} + b_z), z_t \in (-1,1)^h
        \]
        where $W_z \in \mathbb{R}^{h \times d}$, $R_z \in \mathbb{R}^{h \times h}$ are input and recurrent weight matrices, $b_z \in \mathbb{R}^h$ is the bias, and $g(\cdot)$ is typically the $\tanh$ activation function. For an LSTM network, $W$'s, $R$'s and $b$'s denote the input and recurrent weight matrices, and biases of the respective gates with appropriate subscripts.

        \item \textbf{Input gate:}
        \[
        i_t = \sigma(W_i x_t + R_i h_{t-1} + b_i),\, i_t \in (0,1)^h
        \]
        which determines how much of the new information $z_t$ should be written to the memory. Here, $\sigma$ is the sigmoid function and $\odot$ denotes the Hadamard (element-wise) product.

        \item \textbf{Forget gate:}
        \[
        f_t = \sigma(W_f x_t + R_f h_{t-1} + b_f),\, f_t \in (0,1)^h
        \]
        which controls how much of the previous cell state $c_{t-1}$ should be retained.

        \item \textbf{Cell state update:}
        \[
        c_t = i_t \odot z_t + f_t \odot c_{t-1}
        \]
        which combines the new input and old memory state to create the new cell state. The initial cell state is $c_0 = 0$.

        \item \textbf{Output gate:}
        \[
        o_t = \sigma(W_o x_t + R_o h_{t-1} + b_o),\, o_t \in (0,1)^h
        \]
        which decides the parts of the updated memory should be exposed.

        \item \textbf{Hidden state update:}
        \[
        h_t = o_t \odot \sigma_h(c_t)
        \]
        where $\sigma_h(\cdot)$ is typically the $\tanh$ activation and $h_t$ is the hidden state, which is also the output of the LSTM layer. The initial hidden state is $h_0 = 0$.
    \end{itemize}

    \item \textbf{Fully Connected Layer:} The final hidden state $h_T$ (or the sequence $\{h_1, \dots, h_T\}$) is passed to one or more dense layers to map the learned temporal features to the task-specific output space.

    \item \textbf{Output Layer:} The model outputs either a continuous-valued prediction (e.g., electric load estimation) for regression tasks, or a probability distribution over classes for classification tasks.
\end{itemize}

\subsection{Bidirectional Long Short-Term Memory (BiLSTM)}
\label{sec:bilstm}

A Bidirectional LSTM (BiLSTM) network enhances the standard LSTM by merging two parallel LSTM layers: one handles the input sequence in the forward direction, and the other in the backward direction \cite{graves2005framewise}. This architecture facilitates the extraction of contextual information from both past and future time steps, which can be particularly useful for time-dependent signals \cite{huang2015bidirectional} such as periodic electric load data. In the following we review the basics of a BiLSTM network~\cite{graves2005bidirectional}.

\begin{itemize}
    \item \textbf{Input Layer:} The input is a sequence of $d$-dimensional time-ordered vectors: $\{x_1, x_2, \dots, x_T\}$.

    \item \textbf{BiLSTM Layers:} Each time step $t$ is processed by two LSTM layers providing forward and backward pass as explained below. The notation are as defined in subsection \ref{sec:lstm}. The $\rightarrow$ on a vector denotes forward pass and $\leftarrow$ denotes backward pass.

    \begin{itemize}
        \item \textbf{Forward pass:}
        \[
        \overrightarrow{z}_t = g(W_{zf} x_t + R_{zf}\overrightarrow{h}_{t-1} + b_{zf})
        \]
        \[
        \overrightarrow{i}_t = \sigma(W_{if} x_t + R_{if} \overrightarrow{h}_{t-1} +  b_{if})
        \]
        \[
        \overrightarrow{f}_t = \sigma(W_{ff} x_t + R_{ff} \overrightarrow{h}_{t-1} + b_{ff})
        \]
        \[
        \overrightarrow{c}_t = \overrightarrow{i}_t \odot \overrightarrow{z}_t + \overrightarrow{f}_t \odot \overrightarrow{c}_{t-1}
        \]
        \[
        \overrightarrow{o}_t = \sigma(W_{of} x_t + R_{of} \overrightarrow{h}_{t-1} + b_{of})
        \]
        \[
        \overrightarrow{h}_t = \overrightarrow{o}_t \odot \sigma_h(\overrightarrow{c}_t)
        \]

        \item \textbf{Backward pass:}
        \[
        \overleftarrow{z}_t = g(W_{zb} x_t + R_{zb} \overleftarrow{h}_{t+1} + b_{zb})
        \]
        \[
        \overleftarrow{i}_t = \sigma(W_{ib} x_t + R_{ib} \overleftarrow{h}_{t+1} + b_{ib})
        \]
        \[
        \overleftarrow{f}_t = \sigma(W_{fb} x_t + R_{fb} \overleftarrow{h}_{t+1} + b_{fb})
        \]
        \[
        \overleftarrow{c}_t = \overleftarrow{i}_t \odot \overleftarrow{z}_t + \overleftarrow{f}_t \odot \overleftarrow{c}_{t+1}
        \]
        \[
        \overleftarrow{o}_t = \sigma(W_{ob} x_t + R_{ob} \overleftarrow{h}_{t+1} + b_{ob})
        \]
        \[
        \overleftarrow{h}_t = \overleftarrow{o}_t \odot \sigma_h(\overleftarrow{c}_t)
        \]

        \item \textbf{Concatenated output:}
        \[
        h_t = [\overrightarrow{h}_t ; \overleftarrow{h}_t]
        \]
        which represents the full contextual representation at time $t$.
    \end{itemize}

    \item \textbf{Fully Connected Layer:} The concatenated hidden states (e.g., $h_T$ or $\{h_1, h_2, ..., h_T\}$) are flattened and passed to dense layers to learn mappings to the target output.

    \item \textbf{Output Layer:} The final output is a continuous value (e.g., estimated electric load).
\end{itemize}

\subsection{CNN-LSTM}
\label{sec:cnn_lstm}

The CNN-LSTM hybrid model integrates convolutional layers to extract spatial features, followed by LSTM layers for modeling temporal dependencies \cite{shi2015convolutional}. This architecture is well-suited for time-series tasks where input features exhibit both spatial and sequential structure, such as sequences of hourly electric load with weather features \cite{lu2020cnn}. 

\begin{itemize}
    \item \textbf{Input Layer:} The input is a sequence $\{X_1, X_2, \dots, X_T\}$ of matrices (e.g., hourly loads with features), where each $X_t \in \mathbb{R}^{H \times W}$.

    \item \textbf{CNN Layers:} Each input frame $X_t$ is passed through $L$ convolutional layers to extract local spatial features:
    \[
    A_t^{(l)} = \text{ReLU}(W^{(l)} * A_t^{(l-1)} + b^{(l)}), \, 1 \leq l \leq L
    \]
    where $W^{(l)}\in \mathbb{R}^{f_h \times f_w \times K}$ and $b^{(l)}$ are the convolutional kernel and bias at layer $l$, $*$ denotes the convolution operation, and ReLU (Rectified Linear Unit)~\cite{agarap2018deep}  is the activation function. Additionally, $f_h \times f_w$ is the kernel size and $K$ is the number of filters. This process results in a sequence of $d$-dimensional feature maps $\{A_1, A_2, \dots, A_T\}$. The initial input to this is typically $A_t^0 = X_t$.

    \item \textbf{Flattening Layer:} The spatial feature maps $A_t$ are reshaped (flattened) into vectors $a_t \in \mathbb{R}^d$ to form a sequence of feature vectors $\{a_1, a_2, \dots, a_T\}$ that can be input into LSTM.

    \item \textbf{LSTM Layers:} The sequence $\{a_1, \dots, a_T\}$ is passed into an LSTM network to capture temporal dependencies across frames. The LSTM computes hidden states $\{h_1, \dots, h_T\}$ using:
    \[
    h_t = \text{LSTM}(a_t, h_{t-1}, c_{t-1})
    \]
    where $c_{t-1}$ is the cell state from the previous time step. The LSTM layer is as described in Section \ref{sec:lstm}.

    \item \textbf{Fully Connected Layer:} The final LSTM hidden state $h_T$ (or all hidden states) is passed to one or more dense layers for transformation into task-specific features.

    \item \textbf{Output Layer:} Produces the final prediction for regression (e.g., load estimation).
\end{itemize}

\subsection{CNN-BiLSTM}
\label{sec:cnn_bilstm}

The CNN-BiLSTM model enhances the CNN-LSTM architecture by replacing the LSTM with a bidirectional LSTM, which allows the model to train from both past and future dependencies for each time step \cite{lu2021cnn}.

\begin{itemize}
    \item \textbf{Input Layer:} The input is a sequence of spatial frames $\{X_1, X_2, \dots, X_T\}$, where each $X_t$ is a 2D matrix (e.g., hourly loads with features), where each $X_t \in \mathbb{R}^{H \times W}$ for height $H$ and width $W$.

    \item \textbf{CNN Layers:} Each $X_t$ is processed through convolutional layers to obtain spatial feature maps:
    \[
    A_t^{(l)} = \text{ReLU}(W^{(l)} * A_t^{(l-1)} + b^{(l)}), \, 1 \leq l \leq L
    \]
    producing a time-indexed sequence of feature maps. The initial input $A_t^0 = X_t$. The notation of this layer is as described in Section \ref{sec:cnn_lstm}.

    \item \textbf{Flattening Layer:} Each feature map $A_t$ is flattened into a vector $a_t$, forming the sequence $\{a_1, a_2, \dots, a_T\}$.

    \item \textbf{BiLSTM Layers:} The feature sequence is passed through a bidirectional LSTM:
    \[
    \overrightarrow{h}_t = \text{LSTM}_f(a_t, \overrightarrow{h}_{t-1}, \overrightarrow{c}_{t-1}) \quad \text{(forward)}
    \]
    \[
    \overleftarrow{h}_t = \text{LSTM}_b(a_t, \overleftarrow{h}_{t+1}, \overleftarrow{c}_{t+1}) \quad \text{(backward)}
    \]
    \[
    h_t = [\overrightarrow{h}_t ; \overleftarrow{h}_t] \quad \text{(concatenated)}
    \]
    capturing context from both directions. The LSTM layer acts as defined in Section \ref{sec:lstm}. 

    \item \textbf{Fully Connected Layer:} The final (or aggregated) BiLSTM outputs $y_t$ are passed through dense layers to transform into task-specific features.

    \item \textbf{Output Layer:} Produces the output for regression, that is, load estimation.
\end{itemize}

\subsection{Transformer}
\label{sec:transformer}

Transformers are sequence models that rely entirely on self-attention mechanisms \cite{huang2023understanding} rather than recurrence to model dependencies across time steps \cite{vaswani2017attention}. This architecture allows for parallel computation across the sequence and captures global context efficiently.

\begin{itemize}

\item \textbf{Input Layer:}

The input to a Transformer model is a sequence of vectors:
\[
X = \{x_1, x_2, \dots, x_T\},
\]
where $x_t \in \mathbb{R}^d$, \( T \) is the total number of input tokens or time steps, \( d \) is the dimension of each input vector (embedding size), and \( x_t \) is the feature vector for the token at position \( t \). Since Transformers do not have an inherent understanding of sequence order, positional encodings \( P = \{p_1, \dots, p_T\} \) are added to input vectors: $z_t = x_t + p_t$, where \( p_t \in \mathbb{R}^d \) are the positional encoding for time step \( t \), designed to encode relative or absolute position, and \( z_t \in \mathbb{R}^d \) are the position-aware input to the self-attention mechanism.

\item \textbf{Self-Attention Mechanism:}

For each position \( t \), the input \( z_t \) is projected into three different spaces to produce:
\[
Q = z_t W^Q, \quad K = z_t W^K, \quad V = z_t W^V.
\]
Here
\begin{itemize}
    \item \( Q \in \mathbb{R}^{d_k} \): Query vector representing what a token wants to attend to.
    \item \( K \in \mathbb{R}^{d_k} \): Key vector representing what a token contains for others to match against.
    \item \( V \in \mathbb{R}^{d_v} \): Value vector, the actual content or representation to be aggregated.
    \item \( W^Q, W^K, W^V \): Learnable projection matrices of shape \( \mathbb{R}^{d \times d_k} \) or \( \mathbb{R}^{d \times d_v} \).
\end{itemize}

The self-attention output for all positions is computed as:
\[
\text{Attention}(Q, K, V) = \text{softmax}\left( \frac{Q K^\top}{\sqrt{d_k}} \right) V
\]
\begin{itemize}
    \item \( QK^\top \): Measures the correlation between each query and key (attention scores).
    \item \( \sqrt{d_k} \): Scaling factor to prevent overly large dot products (helps stabilize gradients).
    \item \( \text{softmax} \): Converts scores into attention weights summing to 1.
    \item The output is a weighted sum of the value vectors \( V \), where weights depend on similarity between \( Q \) and \( K \).
\end{itemize}

\item \textbf{Multi-Head Attention:}

Transformers replace a single attention mechanism with multiple attention heads running concurrently \cite{cordonnier2020multi} to capture different types of relationships:
\[
\text{MultiHead}(Q, K, V) = [\text{head}_1; \dots; \text{head}_h] W^O
\]
\begin{itemize}
    \item \( h \): Number of attention heads.
    \item \( \text{head}_i = \text{Attention}(Q W_i^Q, K W_i^K, V W_i^V) \): The \( i \)-th head computes attention using its own projection matrices.
    \item \( W_i^Q, W_i^K, W_i^V \): Projection matrices for head \( i \).
    \item \( [\cdot ; \cdot] \): Concatenation of all head outputs.
    \item \( W^O \): Final linear transformation applied after concatenation.
\end{itemize}

By using multiple attention heads, the model can simultaneously capture diverse aspects of the input sequence.

\item \textbf{Feedforward Network:}

After attention, each output vector is passed through a fully connected feedforward neural network:
\[
\text{FFN}(x) = \max(0, x W_1 + b_1) W_2 + b_2
\]
\begin{itemize}
    \item \( x \): Output of attention for a specific position.
    \item \( W_1, W_2 \): Weight matrices for the two linear layers.
    \item \( b_1, b_2 \): Bias terms.
    \item \( \max(0, \cdot) \): ReLU activation function.

\end{itemize}
    \item \textbf{Fully Connected Layer:} The final Transformer encoder (or decoder) output is passed through one or more dense layers to map the high-level contextual embeddings to task-specific features.

    \item \textbf{Output Layer:} Outputs are used for regression, that is, load estimation.
\end{itemize}

\subsection{Prophet}
\label{sec:prophet}

In Prophet, time series forecasting is achieved by decomposing the series into additive elements comprising trend, seasonal patterns, and holiday impacts \cite{jha2021time}. It is robust to missing data and shifts in the trend, and handles outliers well.

\begin{itemize}
    \item \textbf{Model Structure:} Prophet models the time series $y(t)$ as:
    \[
    y(t) = g(t) + s(t) + h(t) + \epsilon_t
    \]
    where $g(t)$ is the trend, $s(t)$ is the seasonality, $h(t)$ captures effects of holidays, and $\epsilon_t$ is the error term.
    
    \item \textbf{Trend Component:} The trend $g(t)$ can be piecewise linear or logistic growth:
    \[
    g(t) = \left(c + (k + a(t)^\top \delta)t \right)\mathbf{I}_{(t \geq s)}
    \]
    where,
    \begin{itemize}
    \item \( g(t) \): The output function evaluated at time \( t \), which is active only for \( t \geq s \).
    
    \item \( c \): A constant term, typically representing a baseline offset or bias.
    
    \item \( k \): A scalar coefficient representing linear growth or decay in the function.
    
    \item \( a(t) \in \mathbb{R}^n \): A time-dependent vector-valued function. It may represent features or basis functions dependent on time \( t \).
    
    \item \( \delta \in \mathbb{R}^n \): A parameter vector to be learned or specified.
    
    \item \( a(t)^\top \delta \): The inner (dot) product between \( a(t) \) and \( \delta \), resulting in a scalar value.
    
    \item \( \left( k + a(t)^\top \delta \right) t \): A time-scaled linear combination of the components \( k \) and \( a(t)^\top \delta \).
    
    \item \( \left( c + \left( k + a(t)^\top \delta \right) t \right) \): The full inner expression which is the main structure of the function when active.
    
    \item \( \mathbf{1}_{(t \geq s)} \): An indicator function defined as:
    \[
    \mathbf{1}_{(t \geq s)} =
    \begin{cases}
    1 & \text{if } t \geq s \\
    0 & \text{otherwise}
    \end{cases}
    \]
    This ensures that \( g(t) \) is non-zero only when \( t \) is greater than or equal to a threshold time \( s \).
\end{itemize}

    \item \textbf{Seasonality Component:} Seasonality is modeled using a Fourier series:
    \[
    s(t) = \sum_{n=1}^{N} \left[a_n \cos\left(\frac{2\pi n t}{P}\right) + b_n \sin\left(\frac{2\pi n t}{P}\right) \right]
    \]
    where $P$ is the period (e.g., yearly, weekly), and $N$ is the number of harmonics.

    \item \textbf{Holiday Component:} Prophet incorporates holiday effects by assigning an indicator function for each holiday with a corresponding parameter.

    \item \textbf{Output:} The forecast is generated by summing the estimated components, providing interpretable forecasts and confidence intervals.
\end{itemize}

\subsection{DLinear}
\label{sec:dlinear}

Dynamic Linear (DLinear) is a simple and effective model for time series forecasting, leveraging decomposition into trend and seasonal parts, each of which is modeled using a linear projection \cite{zeng2023transformers}. It was devised as a simple model for comparing with transformers in time-series forecasting tasks.

\begin{itemize}
    \item \textbf{Input:} A time series sequence $X = \{x_1, x_2, \ldots, x_T\}$, that is, a sequence of electric load data with features appended.

    \item \textbf{Decomposition:} The series is decomposed additively:
    \[
    x_t = s_t + t_t
    \]
    where $s_t$ is the seasonal component and $t_t$ is the trend component.

    \item \textbf{Linear Modeling:} Each component is modeled using separate linear layers:
    \[
    \hat{s} = W_s s + b_s,\quad \hat{t} = W_t t + b_t
    \]
    where $W_s, W_t$ and $b_s, b_t$ are learnable weights and biases.

    \item \textbf{Output:} The forecast is the sum of the predicted components:
    \[
    \hat{x}_{t+h} = \hat{s}_{t+h} + \hat{t}_{t+h}
    \]
\end{itemize}

\subsection{SARIMA}
\label{sec:sarima}

SARIMA (Seasonal AutoRegressive Integrated Moving Average) extends ARIMA (AutoRegressive Integrated Moving Average) by modeling both seasonal and non-seasonal components of a time series \cite{Chen2018time} \cite{box2015time}. It is better suited for electric load forecasting tasks which consists of both seasonal and non-seasonal components.

\begin{itemize}
    \item \textbf{Model Notation:} SARIMA$(p, d, q, P, D, Q)_s$ includes:
    \begin{itemize}
        \item $p$, $d$, $q$: non-seasonal AutoRegressive (AR), differencing and Moving Average (MA) orders respectively
        \item $P$, $D$, $Q$: seasonal AR, differencing and MA orders respectively
        \item $s$: seasonal period
    \end{itemize}

    \item \textbf{Model Equation:}
    \[
    \Phi_P(B^s)\phi_p(B)(1 - B)^d(1 - B^s)^D y_t = \Theta_Q(B^s)\theta_q(B)\epsilon_t
    \]
    where $B$ is the backshift operator \cite{anderson1976backshift}, $\Phi_P$ and $\phi_p$ are the seasonal and non-seasonal AR polynomials \cite{nandi2020data}, $\Theta_Q$ and $\theta_q$ are the seasonal and non-seasonal MA polynomials \cite{shao2015effects}, $\epsilon_t$ is white noise, and $y_t$ is the observed value of the time series at time $t$.

    \item \textbf{Components:}
    \begin{itemize}
        \item AR: AutoRegressive terms based on past values
        \item MA: Moving Average terms based on past errors
        \item Differencing: removes trends and seasonality
    \end{itemize}

    \item \textbf{Output:} Forecasts are computed using the fitted AR and MA equations applied recursively.
\end{itemize}

\subsection{XGBoost}
\label{sec:xgboost}

XGBoost (Extreme Gradient Boosting) is an ensemble tree-based method \cite{ahmad2018tree} that builds a series of decision trees to optimize a given loss function \cite{chen2016xgboost}. Although suited better for classification tasks, XGBoost works surprisingly well in time-series forecasting, especially for electric loads.

\begin{itemize}
    \item \textbf{Objective Function:}
   \begin{equation}
        \mathcal{L} = \sum_{i=1}^{n} l(y_i, \hat{y}_i^{(t)}) + \sum_{k=1}^{t} \Omega(f_k)
    \end{equation}

    where,
    \begin{itemize}
        \item \( \mathcal{L} \): The total objective (loss) function to be minimized.
        \item \( l(y_i, \hat{y}_i^{(t)}) \): A differentiable loss function measuring the prediction error between the true value \( y_i \) and predicted value \( \hat{y}_i^{(t)} \).
        \item \( \Omega(f_k) \): Regularization term which discourages the complexity of tree \( f_k \) and help to prevent overfitting.
        \item \( n \): Number of training instances.
        \item \( t \): Total number of boosting iterations.
    \end{itemize}

    \item \textbf{Model Structure:}
    \begin{equation}
    \hat{y}_i^{(t)} = \sum_{k=1}^{t} f_k(x_i), \quad f_k \in \mathcal{F}
    \end{equation}
    
    where,
    \begin{itemize}
        \item \( \hat{y}_i^{(t)} \): The prediction for sample \( i \) after \( t \) trees have been added.
        \item \( f_k(x_i) \): Output of the \( k \)-th regression tree for input \( x_i \).
        \item \( \mathcal{F} \): The functional space of all possible regression trees.
    \end{itemize}

    \item \textbf{Boosting:} Trees are added sequentially to minimize the loss:
    \begin{equation}
        f_t = \arg \min_f \sum_{i=1}^n \left[ g_i f(x_i) + \frac{1}{2} h_i f(x_i)^2 \right] + \Omega(f)
    \end{equation}
    
    where,
    \begin{itemize}
        \item \( f_t \): The new tree added at iteration \( t \). The new tree $f_t$ is trained to fit the gradients, correcting the previous ensemble’s errors
        \item \( g_i \): First derivative of the loss function with respect to \( \hat{y}_i^{(t-1)} \) (gradient).
        \item \( h_i \): Second derivative (Hessian) of the loss function.
        \item \( f(x_i) \): Prediction of the new tree for sample \( x_i \).
        \item \( \Omega(f) \): Regularization term penalizing the complexity of the new tree.
    \end{itemize}

    \item \textbf{Output:} The final prediction is the sum of the outputs from all trees.
\end{itemize}

\subsection{Random Forest}
\label{sec:rf}

As an ensemble method, Random Forest constructs numerous decision trees and aggregates their predictions to improve the model's ability to generalize~\cite{rigatti2017random}. Like XGBoost (\ref{sec:xgboost}), this model is better suited for classification tasks, yet it works surprisingly well in time-series forecasting, especially for electric loads.

\begin{itemize}
    \item \textbf{Model Structure:} A forest consists of $B$ decision trees:
    \[
    \hat{y} = \frac{1}{B} \sum_{b=1}^{B} f_b(x)
    \]
    where,
    \begin{itemize}
        \item \( \hat{y} \): Final prediction (averaged output).
        \item \( B \): Total number of base models (e.g., decision trees).
        \item \( f_b(x) \): Prediction from the \( b \)-th model for input \( x \).
        \item The summation aggregates predictions from all \( B \) models.
        \item \( \frac{1}{B} \): Averaging the predictions to produce the final output.
    \end{itemize}

    \item \textbf{Tree Diversity:} Each tree is trained with:
    \begin{itemize}
        \item Bootstrap sampling (bagging)
        \item Random feature selection at each split
    \end{itemize}

    \item \textbf{Decision Tree:} Trees are grown to full depth without pruning, and splits are chosen to minimize impurity (e.g., MSE for regression).

    \item \textbf{Output:} For regression tasks, the forest predicts the average of all tree outputs.
\end{itemize}
%
%
%




\putbib
\end{bibunit}


\end{document}